\documentclass{gaussianmerging2k}

\title{Compact Feed-Forward 3D Gaussians\\via Saliency-Guided Primitive Merging}

\addauthor{Tim-Felix Faasch}{tim-felix.faasch@de.bosch.com}{1}
\addauthor{Jochen Kall}{jochen.kall@de.bosch.com}{1}
\addauthor{Cyrill Stachniss}{cyrill.stachniss@igg.uni-bonn.de}{2, 3}

\addinstitution{
  Bosch Research\\
  Hildesheim, Germany
}
\addinstitution{
  University of Bonn\\
  Bonn, Germany
}
\addinstitution{
  Lamarr Institute for Machine Learning and Artificial Intelligence\\
  Bonn, Germany
}

\runninghead{Faasch, Kall, Stachniss}{Compact FF-3DGS via Saliency-Guided Merging}

\def\eg{\emph{e.g}\bmvaOneDot}

\def\ie{\emph{i.e}\bmvaOneDot}

\DeclareMathOperator*{\argmax}{arg\,max}

\usepackage{booktabs}
\usepackage{multirow}
\usepackage{pgfplots}
\pgfplotsset{compat=1.18}
\usepackage{comment}
\usepackage{amsmath,amssymb}
\usepackage{nicefrac}

\def\secref#1{Sec.~\ref{#1}}
\def\figref#1{Fig.~\ref{#1}}
\def\tabref#1{Tab.~\ref{#1}}

\usepackage{tikz}
\usetikzlibrary{positioning, arrows.meta, calc, fit, backgrounds}

\tikzstyle{block}=[rectangle, draw, rounded corners, minimum width=1.5cm,
                   minimum height=1.0cm, align=center, fill=gray!20]
\tikzstyle{cblock}=[rectangle, draw, rounded corners, minimum width=1.5cm,
                    minimum height=1cm, align=center, fill=orange!20]
\tikzstyle{fgnode}=[rectangle, draw, rounded corners, minimum width=1.5cm,
                    minimum height=1cm, align=center, fill=yellow!20]
\tikzstyle{add}=[circle, draw, inner sep=2pt, minimum size=4mm]
\tikzstyle{lbl}=[align=center]

\tikzstyle{imgnode}=[inner sep=1pt, draw=black!20, rounded corners=2pt, line width=0.4pt]

\tikzset{
    flow/.style={->, thick, shorten >=2pt, shorten <=2pt},
    flowU/.style={flow, ultra thick},
    flowNoTip/.style={flow, -},
    flowUNoTip/.style={flowU, -},
    flowDashed/.style={flow, dashed, black!40},
    algoflow/.style={flow, blue!60!black, densely dashed},
    optflow/.style={flow, green!50!black, dotted},
}

\newcommand{\cbox}[5]{%
  \node[#1, #3] (#2) {#4};%
  \def\cboxlbltmp{#5}%
  \ifx\cboxlbltmp\empty
    \node[draw=none, inner sep=0pt, fit=(#2)] (#2_box) {};%
  \else
    \node[font=\scriptsize, below=2pt of #2, inner sep=0pt] (#2_lbl) {#5};%
    \node[draw=none, font=\scriptsize, inner sep=0pt, above=2pt of #2] (#2_top) {\vphantom{#5}};%
    \node[draw=none, inner sep=0pt, fit=(#2)(#2_lbl)(#2_top)] (#2_box) {};%
  \fi
}

\newcommand{\rcanysplatVarFeatvoxelized}{83.2}

\newcommand{\psnranysplatVarFeatmergedkone}{14.14}

\newcommand{\rcanysplatVarFeatmergedktwo}{8.9}
\newcommand{\psnranysplatVarFeatmergedktwo}{14.45}

\newcommand{\psnrdaThreeunmerged}{16.82}

\newcommand{\rcdaThreemergedkone}{4.4}
\newcommand{\psnrdaThreemergedkone}{17.41}

\newcommand{\rcdaThreemergedkfour}{17.3}

\newcommand{\psnrablfeatvariationalsplargemergedkone}{13.97}

\newcommand{\rcablfeatvariationalsplargemergedkfour}{8.0}
\newcommand{\psnrablfeatvariationalsplargemergedkfour}{14.45}

\newcommand{\rcresplatresplat}{6.2}

\newcommand{\rcvolsplatvolsplat}{93.4}

\newcommand{\spCountSmall}{23k}
\newcommand{\spCountMedium}{12k}
\newcommand{\spCountLarge}{5k}

\newcommand{\onlineNaivePerStepM}{1.84}
\newcommand{\onlineKonePerStepM}{0.12}
\newcommand{\onlineSlopeRatioX}{16}
\newcommand{\onlineFinalNaiveM}{12.9}
\newcommand{\onlineFinalKoneM}{0.82}
\newcommand{\onlinePrimRatioX}{16}
\newcommand{\onlineFpsSpeedupX}{6.1}

\newcommand{\psnrdaThreemergedkfourUnmasked}{17.09}
\newcommand{\ssimdaThreemergedkfourUnmasked}{0.488}
\newcommand{\lpipsdaThreemergedkfourUnmasked}{0.520}
\newcommand{\psnrdaThreemergedkoneUnmasked}{16.95}
\newcommand{\ssimdaThreemergedkoneUnmasked}{0.484}
\newcommand{\lpipsdaThreemergedkoneUnmasked}{0.524}
\newcommand{\psnrdaThreemomentmatchingUnmasked}{13.29}
\newcommand{\ssimdaThreemomentmatchingUnmasked}{0.406}
\newcommand{\lpipsdaThreemomentmatchingUnmasked}{0.723}
\newcommand{\psnrdaThreeunmergedUnmasked}{16.13}
\newcommand{\ssimdaThreeunmergedUnmasked}{0.480}
\newcommand{\lpipsdaThreeunmergedUnmasked}{0.430}
\newcommand{\psnrresplatresplatUnmasked}{14.81}
\newcommand{\ssimresplatresplatUnmasked}{0.441}
\newcommand{\lpipsresplatresplatUnmasked}{0.551}
\newcommand{\psnrvolsplatvolsplatUnmasked}{12.24}
\newcommand{\ssimvolsplatvolsplatUnmasked}{0.313}
\newcommand{\lpipsvolsplatvolsplatUnmasked}{0.652}

\newcommand{\rcdaThreemergedkmixed}{9.8}
\newcommand{\psnrdaThreemergedkmixed}{17.55}
\newcommand{\rcdaThreemergedktwoFlexK}{8.5}
\newcommand{\psnrdaThreemergedktwoFlexK}{17.56}

\newcommand{\rcCrossViewRetained}{84.6}

\newcommand{\timingBackboneanysplatVarFeatunmerged}{236}

\newcommand{\timingBackbonedepthsplatunmerged}{443}

\newcommand{\timingBackbonedaThreeunmerged}{387}

\newcommand{\timingTotalresplatresplat}{319}
\newcommand{\timingTotalvolsplatvolsplat}{511}
\newcommand{\bassTimingMs}{13.0}

\newcommand{\bassSegOverheadTwelveViews}{155}

\newcommand{\timingTotalDeployanysplatVarFeat}{765}

\newcommand{\timingTotalDeploydepthsplat}{965}

\newcommand{\timingMergingOverheaddaThree}{540}
\newcommand{\timingTotalDeploydaThree}{925}

\newcommand{\breakEvenFramesThree}{84}

\newcommand{\storageUnmergedMBTwelve}{501}
\newcommand{\storageMergedMBTwelve}{20.6}
\newcommand{\storageCompressionRatioTwelve}{24.4}
\newcommand{\renderMsUnmergedTwelve}{3.83}
\newcommand{\renderMsMergedTwelve}{0.82}
\newcommand{\renderSpeedupTwelve}{4.7}
\newcommand{\mergeOverheadMsTwelve}{452}
\newcommand{\breakEvenFramesTwelve}{150}

\usepackage[textsize=tiny]{todonotes}

\usepackage{acronym}
\providecommand{\acreset}[1]{%
  \expandafter\global\expandafter\let\csname acused@#1@once\endcsname\relax%
  \expandafter\global\expandafter\let\csname acused@#1@twice\endcsname\relax%
}

\makeatletter
\newcommand{\maketitlesupplementary}{%
  \begingroup
    \centering
    {\Large\bfseries \@title\par}%
    \vspace{0.5em}%
    Supplementary Material\par
    \vspace{1.0em}%
  \endgroup
}
\makeatother

\begin{document}

\maketitle

\begin{abstract}
3D scene reconstruction, modeling, and rendering are highly relevant for numerous tasks, and 3D Gaussian splatting has become a standard choice in this context.
Its feed-forward variants provide fast reconstruction from sparse input views but often produce per-pixel primitives, leading to highly redundant and thus inefficient representations.
We present a structure-aware merging pipeline that takes per-pixel primitives from any feed-forward method and consolidates them into a compact, content-adaptive Gaussian set while largely retaining visual quality at just $\nicefrac{1}{20}^\text{th}$ of the Gaussians of a per-pixel method.
We group spatially coherent Gaussians of similar appearance into variable-size clusters via adaptive superpixel segmentation guided by a saliency map, which allocates fine segments to textured regions and coarse segments to homogeneous areas.
We compress each cluster into a compact latent representation through a learned encoder, then match and consolidate representations across views based on geometric overlap and feature similarity via a learned merger.
A level-of-detail decoder then produces the final Gaussians at a controllable resolution, enabling a flexible quality-efficiency trade-off at inference.
As a post-processing module, the pipeline is backbone-agnostic, leveraging the strengths of existing feed-forward methods.
This leads to better and more robust quality than achieved by previous approaches that target a reduction in primitive count, while providing a highly compact representation, that can be rendered efficiently.
\end{abstract}

\acrodef{3DGS}[3DGS]{3D Gaussian Splatting}
\acrodef{FF}[FF]{Feed-Forward}
\acrodef{NeRF}[NeRF]{Neural Radiance Field}
\acrodef{ViT}[ViT]{Vision Transformer}
\acrodef{SH}[SH]{Spherical Harmonics}
\acrodef{EWA}[EWA]{Elliptical Weighted Average}
\acrodef{MLP}[MLP]{Multi-Layer Perceptron}
\acrodef{NVS}[NVS]{Novel View Synthesis}
\acrodef{SSIM}[SSIM]{Structural Similarity Index Measure}
\acrodef{LPIPS}[LPIPS]{Learned Perceptual Image Patch Similarity}
\acrodef{LoD}[LoD]{Level of Detail}

\begin{figure}[h]
\centering
\input{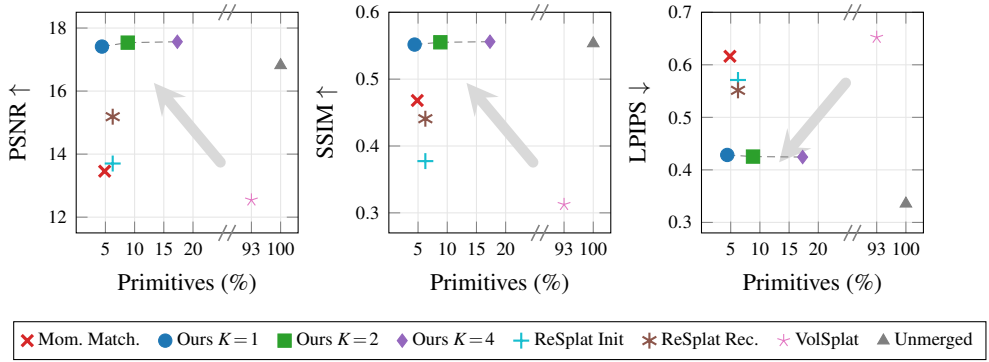}
\caption{NVS quality vs.\ relative primitive count (DA3 backbone, averaged over DL3DV-Bench, MipNeRF360, and Tanks \& Temples). Arrows indicate favorable direction.}
\label{fig:pareto}
\end{figure}

 \section{Introduction}
\label{sec:intro}

3D Gaussian splatting (3DGS)~\cite{Kerbl2023} has emerged as a powerful representation for reconstructing 3D scenes with photorealistic quality, leading to its widespread adoption in simulation~\cite{Xie2024c,Jiang2024}, 3D generation~\cite{Zheng2024,Zhou2024a}, and robotics~\cite{Yan2024,Tonderski2024}.
Traditional per-scene optimization requires dense scene observations and consumes substantial compute, limiting its applicability to large-scale scenarios.
Feed-forward (FF) reconstruction methods address this by predicting Gaussian primitives directly from input views~\cite{Chen2024e,Jiang2025,Ye2024a,Xu2024,Xu2024c,Zhang2024a,Charatan2024,szymanowicz2024splatter}, but they typically produce per-pixel or per-voxel Gaussians, resulting in highly redundant representations~\cite{wang2025learning,Moreau2025}.

Since rendering cost in 3DGS scales with primitive count — each Gaussian must be sorted, projected, and alpha-composited — redundant primitives inflate inference latency and memory footprint without contributing additional visual detail.
This inefficiency is particularly problematic for applications that scale to many scenes and complex downstream tasks, such as simulation-driven training of models for robotics or autonomous driving, where fast rendering is crucial for large-scale data generation.
Redundancy can also hinder generalization to novel views, as overfitting to input views may lead to high-frequency artifacts.
To address these challenges, we focus on reducing the number of primitives in FF-reconstructed scenes while maintaining visual quality.
A good merging strategy should be content-adaptive, allocating representational capacity where it matters most, and it should be able to consolidate redundant primitives across overlapping views.

Existing approaches have made significant progress in addressing the problem of redundant Gaussians through spatial discretization~\cite{wang2025volsplat,Jiang2025,wang2025learning}, iterative refinement~\cite{Xu2025a}, or through detection-based primitive placement~\cite{Moreau2025}.
While these methods achieve notable reductions in primitive count with good rendering quality, opportunities for further improvement remain: uniform discretization strategies do not fully exploit the varying visual importance across scene regions, iterative refinement methods can lead to overfitting on input views, and detection-based methods operate during initial reconstruction without support for merging primitives across multiple views or adapting to new reconstruction backbones.
We aim to address these open challenges with our work.

The main contribution of this paper is a novel technique to substantially reduce the complexity of 3D Gaussian scenes that feed-forward methods produce.
We propose a structure-aware, superpixel-based primitive merging strategy that takes per-pixel Gaussians from any feed-forward method and consolidates them into a highly compact representation.
Content-adaptive superpixel segmentation, guided by local image structure, groups spatially coherent, perceptually similar per-pixel Gaussians.
A learned set-attention encoder compresses each group into a compact latent representation, which we then match across views and merge using a learned fusion module.
A local self-attention module refines the result, and a decoder produces output Gaussians at a controllable level of detail.
This encode-merge-decode pipeline retains the bulk of the novel view synthesis quality at just $\nicefrac{1}{20}^\text{th}$ of the Gaussians of a per-pixel method, vastly increasing the rendering speed for downstream applications while providing more robust and reliable reconstructions than competing reduced-primitive methods, particularly in sparse-view settings.
Crucially, the merging module attaches to any per-pixel Gaussian predictor without retraining the backbone, making it a flexible solution for improving the efficiency of existing FF 3DGS methods.

\section{Related Work}
\label{sec:related}

\textbf{3D Scene Representations:}
Traditional computer graphics has long relied on explicit representations such as meshes and point clouds for rasterization-based rendering pipelines~\cite{AkenineMoeller2018}.
Neural Radiance Fields (NeRF)~\cite{Mildenhall2020} introduced a neural implicit approach that learns to represent scenes from multi-view imagery, enabling applications in reconstruction~\cite{Mildenhall2020,Barron2022}, simulation~\cite{Tonderski2024}, and content generation~\cite{Sargent2023,Qian2023}.
Despite their quality, NeRFs suffer from high computational demands due to volumetric rendering.
3D Gaussian splatting (3DGS)~\cite{Kerbl2023} is an explicit representation that addresses this limitation by representing scenes as collections of 3D Gaussians rendered through differentiable rasterization, achieving both faster optimization and interactive rendering frame rates.
However, 3DGS still requires per-scene optimization, which incurs significant computational cost when scaling to a large number of scenes and fails to converge to a meaningful geometry for sparse-view settings.

\textbf{Feed-Forward Reconstruction:}
To overcome the limitations of per-scene optimization, feed-forward (FF) methods predict 3D Gaussians directly from input views~\cite{Chen2024e,Jiang2025,Ye2024a,Xu2024,Xu2024c,Zhang2024a,Charatan2024,szymanowicz2024splatter}.
The backbone is typically a large vision transformer (ViT)~\cite{dosovitskiy2020image} that encodes the input views and regresses the Gaussian parameters.
FF reconstruction runs significantly faster than per-scene optimization and can handle sparse-view settings, but often places Gaussians less efficiently.
Most existing FF methods predict per-pixel Gaussians~\cite{Charatan2024,szymanowicz2024splatter,Chen2024e,Xu2024,Xu2024c,Ye2024a,Zhang2024a,Lin2025a}, which results in a highly redundant representation with an excessive number of sub-optimally placed primitives.

\textbf{Efficient 3D Gaussian Splatting:}
The number of primitives in a 3DGS reconstruction directly affects rendering speed, memory consumption, and storage cost.
Several strategies aim to reduce this overhead.
Compression techniques such as quantization and learned entropy coding lower the per-primitive memory footprint without reducing the number of Gaussians~\cite{wang2025smol,Chen2024f,Lee2024a,Niedermayr2024,Chen2024HAC,hac++2025}.
In contrast, compaction methods aim to reduce the primitive count itself: hierarchical representations merge primitives that contribute primarily to the background~\cite{kerbl2024hierarchical}, and modified densification strategies prune redundant Gaussians during optimization~\cite{mallick2024taming,Fang2024,Lee2025a,Bai2025,Lee2024a}.
Compression and compaction are complementary: attribute-compression schemes such as HAC/HAC++~\cite{Chen2024HAC,hac++2025} can be applied on compact Gaussian scenes for further storage savings.
However, both compression and compaction methods of this kind are tightly coupled to the per-scene optimization process, and are therefore not directly compatible with feed-forward reconstruction methods.

Some methods specifically tackle the efficiency of FF-produced representations.
The most common approach is voxelization, where some methods directly predict voxel-aligned Gaussians~\cite{wang2025volsplat,Miao2025,Ren2024a} and others apply post-processing voxelization~\cite{Jiang2025,wang2025learning}; yet uniform spatial discretization does not account for varying visual complexity across scene regions.
Detection-based approaches~\cite{Moreau2025} move from per-pixel regression to importance-driven Gaussian placement, allocating primitives according to visual saliency.
Iterative refinement methods~\cite{Xu2025a} predict Gaussians in a subsampled space and refine them using gradient-free feedback from the rendering error, yielding $16\times$ fewer primitives than per-pixel methods.
Both directions integrate Gaussian reduction directly into the reconstruction model itself, which is fundamentally different and complementary to our post-hoc compression approach.
While these methods potentially allow for more efficient FF reconstruction, they are less flexible than our post-processing pipeline, and cannot directly leverage the strong performance of existing FF backbones.
Graph-based fusion mechanisms~\cite{Wang2024h,Zhang2024c} address the redundancy of Gaussians in overlapping views by progressively merging per-pixel Gaussians across views via depth-proximity matching, GRU-based feature updates, or graph convolutions with overlap-weighted edges followed by pooling layers that prune geometrically close primitives.
These methods rely on globally uniform proximity thresholds and do not account for visual complexity, allocating primitives uniformly regardless of scene content.
In contrast, our method focuses on redistributing primitive capacity based on visual complexity, in addition to addressing the redundancy of overlapping views, which allows for much greater compression of the Gaussian representation while retaining high reconstruction quality.
A complementary direction compresses redundant multi-view inputs into a compact latent state before Gaussian prediction using an Information Bottleneck formulation, enabling FF models to scale to over a hundred input views~\cite{wang2026zpressor}.
This acts as an input-side compression module and does not address the redundancy of the predicted per-pixel Gaussians; it could therefore be combined with our output-side primitive merging to obtain an end-to-end efficient pipeline. 

\textbf{Superpixel Segmentation:}
Superpixel segmentation is a widely used technique in computer vision that groups pixels into small, perceptually meaningful regions based on color and spatial proximity~\cite{Achanta2010,868688,zhang2011superpixels}. 
Practitioners often apply it as a pre-processing step for tasks such as object recognition, image segmentation, and scene understanding to reduce computational complexity and improve efficiency of subsequent processing steps \cite{barcelos2024comprehensive}. 
Recent superpixel methods with adaptive sizing capabilities~\cite{uziel2019bayesian,zhao2025content} are particularly effective, as they can allocate more segments to visually complex regions.
However, as far as we are aware, no prior work has applied these methods to 3D scene reconstruction.

\section{Preliminaries}
\label{sec:prelim}

\textbf{3D Gaussian Splatting.}
In 3DGS~\cite{Kerbl2023}, a scene is represented as a set of $N$ anisotropic Gaussians $\mathcal{G} = \{\mathcal{G}_1, \dots, \mathcal{G}_N\}$.
Each Gaussian~$\mathcal{G}_i$ is parameterized by a mean $\boldsymbol{\mu}_i \in \mathbb{R}^3$ and a covariance $\boldsymbol{\Sigma}_i \in \mathbb{R}^{3 \times 3}$, with its evaluation at position $\mathbf{x} \in \mathbb{R}^3$ given by
\begin{equation}
\label{eq:gaussian}
\mathcal{G}_i(\mathbf{x}) = \exp\!\left(-\tfrac{1}{2}(\mathbf{x} - \boldsymbol{\mu}_i)^\top \boldsymbol{\Sigma}_i^{-1} (\mathbf{x} - \boldsymbol{\mu}_i)\right).
\end{equation}
For rendering, Gaussians are projected into image space via EWA splatting~\cite{Zwicker2001}, yielding a 2D covariance $\boldsymbol{\Sigma}_i' = \mathbf{J}\mathbf{W}\boldsymbol{\Sigma}_i\mathbf{W}^\top\mathbf{J}^\top$ with the viewing transformation~$\mathbf{W}$ and the projection Jacobian~$\mathbf{J}$.
The 3D covariance is decomposed as $\boldsymbol{\Sigma}_i = \mathbf{R}_i \mathbf{S}_i \mathbf{S}_i^\top \mathbf{R}_i^\top$ into rotation~$\mathbf{R}_i$ and diagonal scale~$\mathbf{S}_i$, and is parameterized by a quaternion $\mathbf{q}_i \in \mathbb{R}^4$ and scale vector $\mathbf{s}_i \in \mathbb{R}^3$.
The view-dependent color $\mathbf{c}_i$ of $\mathcal{G}_i$ is encoded via spherical harmonic (SH) coefficients $\hat{\mathbf{c}}_{0,i}, \dots, \hat{\mathbf{c}}_{M,i}$, where $\mathbf{c}_i = \sum_{k=0}^{M} \hat{\mathbf{c}}_{k,i}\, H_k(\mathbf{d})$ for view direction~$\mathbf{d} \in \mathbb{R}^3$, with the SH basis functions~$H_k$.
The final pixel color is obtained by front-to-back alpha blending:
\begin{equation}
\label{eq:alpha_blending}
\mathbf{c}_\text{pixel} = \sum_{i=1}^{N} T_i \, \alpha_i \, \mathbf{c}_i, \quad T_i = \prod_{j=1}^{i-1}(1 - \alpha_j),
\end{equation}
where $T_i$ is the transmittance, and $\alpha_i$ the opacity of $\mathcal{G}_i$.
Each Gaussian is fully described by its set of parameters $\Theta_i = (\boldsymbol{\mu}_i, \mathbf{q}_i, \mathbf{s}_i, \alpha_i, \hat{\mathbf{c}}_{0 \dots M, i})$.
In the standard per-scene optimization setting, these parameters are optimized via gradient descent, minimizing a combination of different photometric losses between images rendered through differentiable rasterization and ground-truth images.

\textbf{Feed-Forward Gaussian Splatting.}
Feed-forward (FF) methods~\cite{Chen2024e,Jiang2025,Xu2024c,Lin2025a} bypass per-scene optimization by predicting Gaussian parameters directly from $N$ input views $\{\mathbf{I}_v\}_{v=1}^N$. 
A typical architecture feeds each view through a vision transformer (ViT)~\cite{dosovitskiy2020image} backbone to extract per-patch features, and applies multi-view cross-attention to obtain multi-view-aware feature maps. 
Prediction heads then regress point maps, which can be projected into 3D space to retrieve the Gaussian means $\boldsymbol{\mu}_i$ using known camera parameters.
The remaining parameters are similarly predicted from the patch features.
FF methods have varying input requirements: \emph{posed} methods require known camera intrinsics and extrinsics \cite{Charatan2024, Chen2024e}, whereas \emph{unposed} methods jointly predict camera parameters and scene geometry \cite{Ye2024a, Jiang2025}.

\section{Saliency-Guided Superpixel Merging}
\label{sec:method}

\begin{figure}
    \input{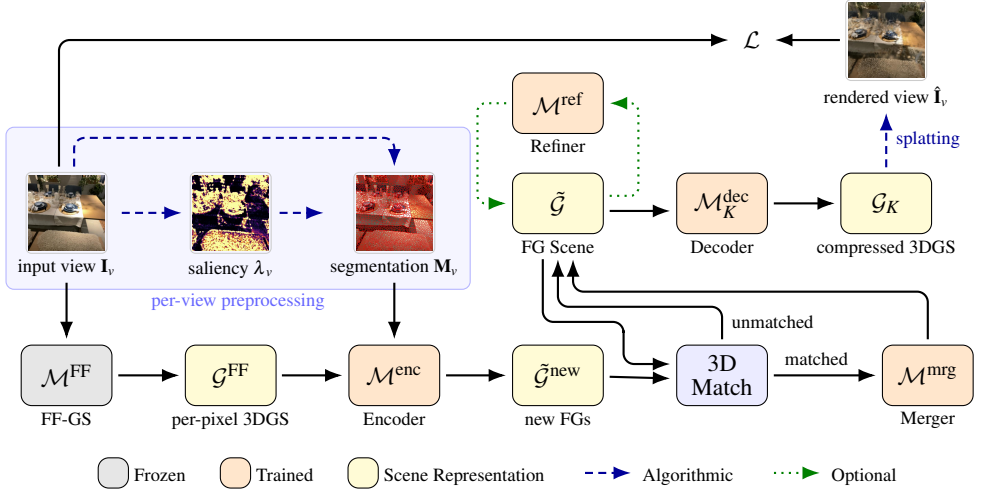}
    \caption{Pipeline overview.
    Per-pixel Gaussians~$\mathcal{G}^\text{FF}$ are grouped into saliency-guided superpixels.
    The encoder~$\mathcal{M}^\text{enc}$ compresses each group into a single Feature Gaussian (FG), the merger~$\mathcal{M}^\text{mrg}$ fuses FGs that overlap across views, the refiner~$\mathcal{M}^\text{ref}$ updates each FG based on its neighbors.
    The decoder~$\mathcal{M}^\text{dec}_K$ expands each FG into $K$ output Gaussians, resulting in the output scene~$\mathcal{G}_K$.}
    \vspace{-0.4cm}
    \label{fig:method}
\end{figure}

Given per-pixel Gaussian primitives predicted by an arbitrary frozen feed-forward backbone $\mathcal{M}^\text{FF}$, our method consolidates them into a compact representation that largely retains visual fidelity.
A key design goal is content-adaptivity: homogeneous regions (\eg, sky, walls) can be aggressively compressed, whereas fine detail (\eg, at edges or textures) requires a higher representational capacity.
We achieve this through an encode-merge-decode pipeline depicted in~\figref{fig:method}.
First, saliency-guided superpixel segmentation groups spatially coherent, perceptually similar Gaussians at a content-adaptive granularity (\secref{sec:grouping}).
A learned encoder~$\mathcal{M}^\text{enc}$ compresses each group into a latent-augmented representation we call a \emph{Feature Gaussian} (\secref{sec:encoder}).
A level-of-detail decoder $\mathcal{M}^\text{dec}_K$ reconstructs 3D Gaussians, compatible with existing rendering pipelines, at a controllable resolution $K$ (\secref{sec:decoder}).
Cross-view matching identifies overlapping Feature Gaussians, and a learned merger~$\mathcal{M}^\text{mrg}$ consolidates them, enabling further compression (\secref{sec:merging}).
A local self-attention refiner~$\mathcal{M}^\text{ref}$ provides neighborhood awareness so that thin structures spanning multiple groups can be faithfully reconstructed (\secref{sec:refinement_online}).
The pipeline is trained end-to-end with photometric losses on the rendered output (\secref{sec:training}).

\subsection{Content-Adaptive Superpixel Grouping}
\label{sec:grouping}

The first step partitions each view's per-pixel Gaussians into groups that the encoder can compress.
A good group for merging should be color-homogeneous (so appearance can be preserved), spatially contiguous (necessary for 3D coherence), and have regular shape (easier to represent with few Gaussians).
At the same time, group size should adapt to local scene complexity: edges and corners carry geometric detail that merging would destroy, so they require smaller groups and thus more output primitives; flat regions can be aggressively merged.

\begin{figure*}[t]
    \centering
    \subfigure[Input $\mathbf{I}_v$]{%
        \includegraphics[width=0.185\linewidth]{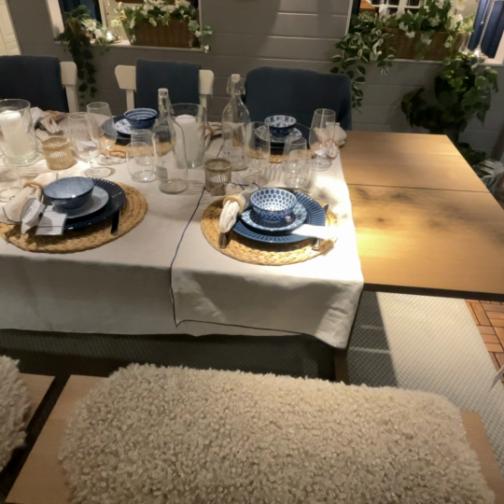}}%
    \hfill%
    \subfigure[Saliency $\boldsymbol{\lambda}_v$]{%
        \includegraphics[width=0.185\linewidth]{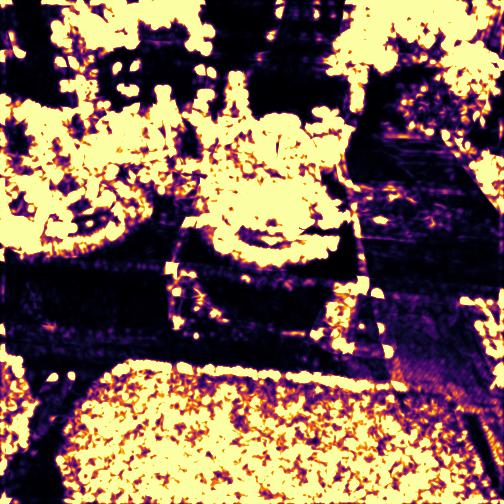}}%
    \hfill%
    \subfigure[SLIC]{%
        \includegraphics[width=0.185\linewidth]{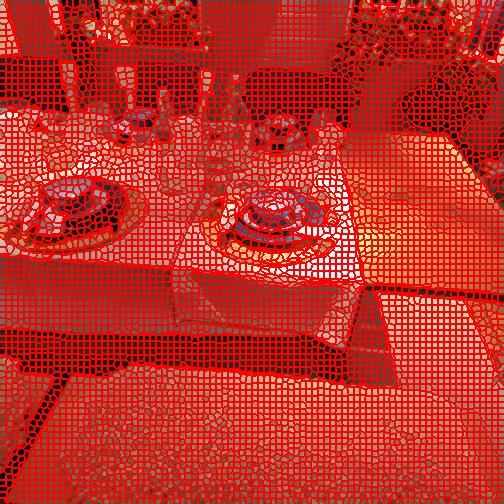}}%
    \hfill%
    \subfigure[BASS]{%
        \includegraphics[width=0.185\linewidth]{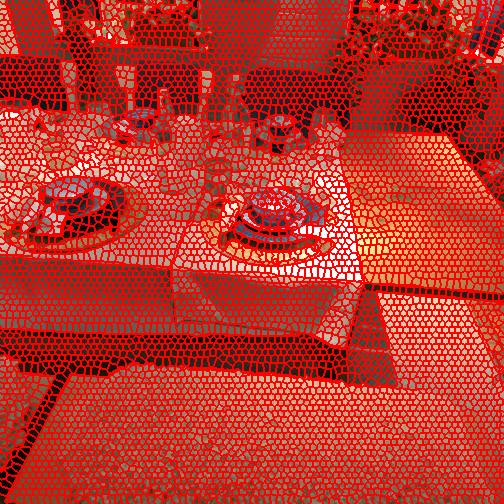}}%
    \hfill%
    \subfigure[BASS $+\boldsymbol{\lambda}_v$]{%
        \includegraphics[width=0.185\linewidth]{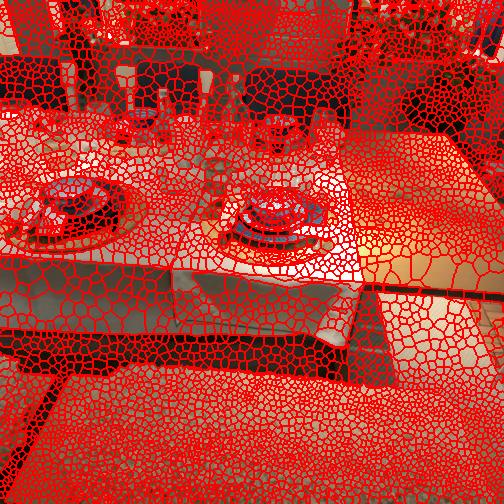}}%
    \caption{Superpixel comparison at similar segment count: SLIC (uniform), BASS with uniform seeds, and BASS with saliency-guided seeds~$\boldsymbol{\lambda}_v$.}
    \label{fig:segmentation}
    \vspace{-0.25cm}
\end{figure*}

Bayesian Adaptive Superpixel Segmentation (BASS)~\cite{uziel2019bayesian} produces segments with the first three properties by iteratively maximizing a posterior that balances color homogeneity with spatial compactness, while maintaining regular segment boundaries.
Standard BASS initializes seeds on a uniform hexagonal grid, and achieves content adaptivity by iteratively shifting superpixel centers and borders.
For very small segment sizes, this can still lead to a large number of segments in flat regions, which is undesirable for our application.
We replace the uniform initialization with a saliency-guided seed placement: we compute the Shi-Tomasi corner response~$\boldsymbol{\lambda}_v$~\cite{Shi1994} at each pixel (the minimum eigenvalue of the image structure tensor~\cite{scharr2000optimale}) and sample seeds densely where this response is high and sparsely elsewhere.
BASS then refines these seeds into final segments of similar color and a regular shape.
The output is a per-view partition map $\mathbf{M}_v$ that assigns each pixel (and its associated Gaussian) to a superpixel group $\mathcal{S}_j$.
Through the combination of saliency-guided seeding and BASS segmentation, we obtain small segments in regions of high detail and larger segments in flatter regions, effectively adapting to local scene complexity (\figref{fig:segmentation}).

\subsection{Feature Gaussian Encoder}
\label{sec:encoder}

\begin{figure}[t]
    \centering
    \def\modBlkW{0.8cm}
\def\modBlkH{0.5cm}
\def\modColGap{0.5cm}
\def\modRowGap{0.8cm}   %
\def\modFont{\scriptsize}
\def\modLblFont{\tiny}
\def\tokenSep{0.2cm}   %

\begin{tikzpicture}[
  font=\modFont,
  >=Latex,
  node distance=0.5cm and \modColGap,
  modblock/.style={rectangle, draw, rounded corners=2pt, minimum width=\modBlkW,
                   minimum height=\modBlkH, align=center, fill=orange!20,
                   font=\modFont},
  datanode/.style={rectangle, draw, rounded corners=2pt, minimum width=0.75cm,
                   minimum height=\modBlkH, align=center, fill=yellow!15,
                   font=\modFont},
  tokennode/.style={rectangle, draw, rounded corners=1pt, minimum width=0.42cm,
                    minimum height=\modBlkH, fill=yellow!15, inner sep=1pt, align=center,
                    font=\modFont},
  modarrow/.style={->, thick, shorten >=1pt, shorten <=1pt, rounded corners=1pt},
  sabflow/.style={->, thin, shorten >=1pt, shorten <=1pt, gray!60},
  sublabel/.style={font=\modLblFont, text=black!60},
  grouplabel/.style={font=\small\bfseries, anchor=west},
]

\node[grouplabel, anchor=north west] (enc_title) {\small(a) Encoder $\mathcal{M}^\text{enc}$ / Merger $\mathcal{M}^\text{mrg}$};

\node[tokennode, below=0.5cm of enc_title.west, xshift=0.5cm, anchor=north] (enc_tok1) {$\mathcal{G}_1$};
\node[tokennode, below=\tokenSep of enc_tok1] (enc_tok2) {$\mathcal{G}_2$};
\node[below=1pt of enc_tok2, sublabel] (enc_tokdots) {$\vdots$};
\node[tokennode, below=\tokenSep of enc_tokdots] (enc_tokn) {$\mathcal{G}_n$};
\node[sublabel, below=1pt of enc_tokn] (enc_in_lbl) {$n$ inputs};

\node[modblock, right=\modColGap of enc_tok1] (enc_mlp1) {MLP};
\node[modblock, right=\modColGap of enc_tok2] (enc_mlp2) {MLP};
\node[below=1pt of enc_mlp2, sublabel] (enc_mlpdots) {$\vdots$};
\node[modblock, right=\modColGap of enc_tokn] (enc_mlpn) {MLP};
\node[sublabel, below=1pt of enc_mlpn] (enc_enc_lbl) {encode};

\node[right=\modColGap of enc_mlp1, anchor=west] (sab1) {};
\node[right=\modColGap of enc_mlp2, anchor=west] (sab2) {};
\node[right=\modColGap of enc_mlpn, anchor=west] (sabn) {};
\def\sabgap{0.1cm}
\node[right=\sabgap of sab1, anchor=west] (sab_span1) {};
\node[right=\sabgap of sabn, anchor=west] (sab_span2) {};
\node[modblock, fit=(sab_span1)(sab_span2)] (enc_sab) {};
\node at (enc_sab) {SAB};
\node[sublabel, below=1pt of enc_sab] {self-attn};

\node[modblock, anchor=center,
      yshift=0pt] (enc_pma) at ([xshift=1cm]enc_sab.east |- enc_sab.center) {PMA};
\node[tokennode, above=0.5cm of enc_pma] (seed_pma) {$\mathcal{Q}$};

\node[modblock] (enc_heads) at (enc_pma |- enc_tokn) {MLP};
\node[sublabel, below=1pt of enc_heads] {aggregate \& predict};

\node[tokennode, right=\modColGap of enc_heads] (enc_out) {$\tilde{\mathcal{G}}$};
\node[sublabel, below=1pt of enc_out] (enc_out_lbl) {1 output};

\draw[modarrow] (enc_tok1.east) -- (enc_mlp1.west);
\draw[modarrow] (enc_tok2.east) -- (enc_mlp2.west);
\draw[modarrow] (enc_tokn.east) -- (enc_mlpn.west);

\draw[sabflow] (enc_mlp1.east) -- (sab2.west);
\draw[sabflow] (enc_mlp1.east) -- (sabn.west);
\draw[sabflow] (enc_mlp2.east) -- (sab1.west);
\draw[sabflow] (enc_mlp2.east) -- (sabn.west);
\draw[sabflow] (enc_mlpn.east) -- (sab1.west);
\draw[sabflow] (enc_mlpn.east) -- (sab2.west);

\draw[modarrow] (enc_mlp1.east) -- (enc_mlp1.east -| enc_sab.west);
\draw[modarrow] (enc_mlp2.east) -- (enc_mlp2.east -| enc_sab.west);
\draw[modarrow] (enc_mlpn.east) -- (enc_mlpn.east -| enc_sab.west);

\draw[modarrow] (enc_sab.east |- enc_mlp1.center) -- ([xshift=-0.4cm]enc_pma.west |- enc_mlp1.center)
    |- (enc_pma.west);
\draw[modarrow] (enc_sab.east |- enc_mlp2.center) -- ([xshift=-0.4cm]enc_pma.west |- enc_mlp2.center)
    |- (enc_pma.west);
\draw[modarrow] (enc_sab.east |- enc_mlpn.center) -- ([xshift=-0.4cm]enc_pma.west |- enc_mlpn.center)
    |- (enc_pma.west);

\draw[modarrow] (seed_pma.south) -- (enc_pma.north);

\draw[modarrow] (enc_pma.south) -- (enc_heads.north);
\draw[modarrow] (enc_heads.east) -- (enc_out.west);

\node[grouplabel, anchor=north west] at (6.5, 0) (dec_title) {\small(b) Decoder $\mathcal{M}^\text{dec}_K$};

\node[tokennode, below=0.5cm of dec_title.west, xshift=0.5cm, anchor=north] (dec_in) {$\tilde{\mathcal{G}}$};
\node[sublabel, below=1pt of dec_in] (dec_in_lbl) {1 input};

\node[modblock, right=0.7cm of dec_in] (dec_slot_1) {MLP$_1$};
\node[modblock, below=\tokenSep of dec_slot_1] (dec_slot_2) {MLP$_2$};
\node[below=1pt of dec_slot_2, sublabel] (dec_indots) {$\vdots$};
\node[modblock] (dec_slot_K) at (dec_slot_1 |- enc_tokn) {MLP$_K$};
\node[sublabel, below=1pt of dec_slot_K] {$K$ slots};

\node[right=\modColGap of dec_slot_1, anchor=west] (dec_sab1) {};
\node[right=\modColGap of dec_slot_2, anchor=west] (dec_sab2) {};
\node[right=\modColGap of dec_slot_K, anchor=west] (dec_sabK) {};
\def\decSabGap{0.1cm}
\node[right=\decSabGap of dec_sab1, anchor=west] (dec_sab_span1) {};
\node[right=\decSabGap of dec_sabK, anchor=west] (dec_sab_span2) {};
\node[modblock, fit=(dec_sab_span1)(dec_sab_span2)] (dec_sab) {};
\node at (dec_sab) (dec_sab_lbl) {SAB};
\node[sublabel, below=1pt of dec_sab] {self-attn};

\node[modblock] (dec_head1) at ([xshift=2*\modColGap]dec_sab.east|-dec_slot_1.center) {MLP};
\node[modblock] (dec_head2) at ([xshift=2*\modColGap]dec_sab.east|-dec_slot_2.center) {MLP};
\node[below=1pt of dec_head2, sublabel] (dec_mlpdots) {$\vdots$};
\node[modblock] (dec_headK) at ([xshift=2*\modColGap]dec_sab.east|-dec_slot_K.center) {MLP};
\node[sublabel, below=1pt of dec_headK] {predict};

\node[tokennode, right=\modColGap of dec_head1] (dec_out1) {$\mathcal{G}_1$};
\node[tokennode, right=\modColGap of dec_head2] (dec_out2) {$\mathcal{G}_2$};
\node[below=1pt of dec_out2, sublabel] (dec_tokdots) {$\vdots$};
\node[tokennode, right=\modColGap of dec_headK] (dec_outK) {$\mathcal{G}_K$};
\node[sublabel, below=1pt of dec_outK] (dec_out_lbl) {$K$ outputs};

\draw[modarrow] (dec_in.east) -- (dec_slot_1.west);
\draw[modarrow] (dec_in.east) -- ++(0.25, 0) |- (dec_slot_2.west);
\draw[modarrow] (dec_in.east) -- ++(0.25, 0) |- (dec_slot_K.west);

\draw[sabflow] (dec_slot_1.east) -- (dec_sab2.west);
\draw[sabflow] (dec_slot_1.east) -- (dec_sabK.west);
\draw[sabflow] (dec_slot_2.east) -- (dec_sab1.west);
\draw[sabflow] (dec_slot_2.east) -- (dec_sabK.west);
\draw[sabflow] (dec_slot_K.east) -- (dec_sab1.west);
\draw[sabflow] (dec_slot_K.east) -- (dec_sab2.west);

\draw[modarrow] (dec_slot_1.east) -- (dec_slot_1.east -| dec_sab.west);
\draw[modarrow] (dec_slot_2.east) -- (dec_slot_2.east -| dec_sab.west);
\draw[modarrow] (dec_slot_K.east) -- (dec_slot_K.east -| dec_sab.west);

\draw[modarrow] (dec_sab.east |- dec_slot_1.center) -- (dec_head1.west);
\draw[modarrow] (dec_sab.east |- dec_slot_2.center) -- (dec_head2.west);
\draw[modarrow] (dec_sab.east |- dec_slot_K.center) -- (dec_headK.west);

\draw[modarrow] (dec_head1.east) -- (dec_out1.west);
\draw[modarrow] (dec_head2.east) -- (dec_out2.west);
\draw[modarrow] (dec_headK.east) -- (dec_outK.west);

\begin{scope}[on background layer]
  \node[draw=orange!40, fill=orange!4, rounded corners=1pt, inner sep=2pt,
        fit=(enc_title)(enc_enc_lbl)(enc_out_lbl)] (enc_bg) {};
  \node[draw=orange!40, fill=orange!4, rounded corners=1pt, inner sep=2pt,
        fit=(dec_title)(dec_in_lbl)(dec_out_lbl)] (dec_bg) {};
\end{scope}

\end{tikzpicture}

\vspace{-0.15cm}
    \caption{Internal architecture of our modules built on the Set Transformer~\cite{Lee2019} building blocks SAB and PMA.
    (a)~The encoder and merger share the same many-to-one pattern: $n$ input tokens are refined by SABs and aggregated into a single output via PMA and a learnable query $\mathcal{Q}$.
    (b)~The decoder inverts this pattern: a single Feature Gaussian is expanded into $K$ output Gaussians via learned slot tokens.}
    \label{fig:architecture}
\end{figure}
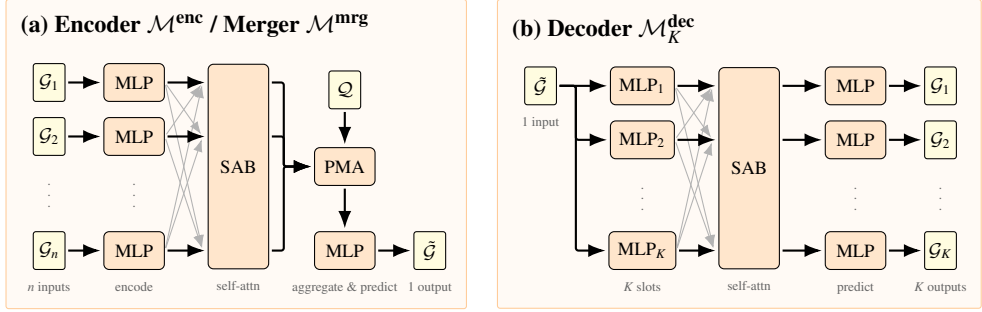

Given a superpixel group $\mathcal{S}_j$, the encoder $\mathcal{M}^\text{enc}$ compresses its Gaussians into an intermediate representation that downstream stages (matching, merging, refinement) can process efficiently.
To achieve this, we introduce the \emph{Feature Gaussian}~$\tilde{\mathcal{G}}_j$, a latent-augmented representation that encodes the entire group with a single set of parameters.
A Feature Gaussian has the same spatial format as a regular Gaussian (position~$\boldsymbol{\mu}_j$, scale~$\mathbf{s}_j$, and rotation~$\mathbf{q}_j$), but replaces view-dependent color with a base color~$\mathbf{c}_j$ and a latent vector~$\mathbf{z}_j \in \mathbb{R}^d$.
The latent stores the visual appearance and geometric structure of the entire group so that the decoder can later expand it into one or more output Gaussians that capture the group's complexity.

Compressing a variable-size set of Gaussians into a single Feature Gaussian requires an architecture that is permutation-invariant (the output should not depend on input order) and handles arbitrary group sizes.
We use the Set Transformer~\cite{Lee2019}: an MLP projects per-Gaussian attributes into tokens, a stack of Set Attention Blocks (SABs) refines them via pairwise interactions, and a Pooling-by-Multihead-Attention (PMA) layer aggregates the set into a single output using a learnable query $\mathcal{Q}$ (\figref{fig:architecture}).
We bound the encoder's compute budget for very large superpixels as detailed in the supplementary material (\secref{sec:encoder_compute}).
A final MLP produces the latent feature~$\mathbf{z}_j$, and a position offset from the opacity-weighted centroid, allowing the encoder to compensate for outliers that would otherwise skew the placement of the Feature Gaussian.
The base color~$\mathbf{c}_j$ is the opacity-weighted average of the group members' diffuse color.
Notably, the encoder does not predict the scale~$\mathbf{s}_j$ or rotation~$\mathbf{q}_j$ that are needed for cross-view matching; these are produced by the decoder $\mathcal{M}^\text{dec}_1$.
This design prevents a collapse mode where the encoder shrinks geometry to avoid cross-view matching entirely, which would simplify reconstruction but hurts compression.

\subsection{Level-of-Detail Decoder}
\label{sec:decoder}

The decoder $\mathcal{M}^\text{dec}_K$ produces the $K$ output Gaussians from each Feature Gaussian, which are compatible with any differentiable rasterizer and can be directly rendered.
We train multiple decoder heads for different $K$: the $K\!=\!1$ head provides the geometry of the Feature Gaussian, which is used for cross-view matching and serves as maximum-compression output, while heads with $K\!>\!1$ trade a higher number of Gaussians for higher reconstruction fidelity.
This exposes $K$ as an inference-time knob, allowing the user to balance quality against rendering compute cost without retraining.

Expanding a single latent into $K$ outputs requires learned specialization.
We use a slot-based design (\figref{fig:architecture}): a slot-seed MLP maps the latent feature~$\mathbf{z}_j$ and base color~$\mathbf{c}_j$ into $K$ slot tokens, per-slot self-attention lets slots coordinate, and output heads predict each Gaussian's parameters.
Mean~$\boldsymbol{\mu}_i$ and base color~$\mathbf{c}_i$ are predicted as offsets from the Feature Gaussian's parameters.
At inference, we prune Gaussians with $\alpha_i < \tau_a$ for further compression.

\subsection{Cross-View Matching and Merging}
\label{sec:merging}

Feature Gaussians~$\tilde{\mathcal{G}}$ from different views that observe the same 3D region should be merged to reduce redundancy.
We identify candidate groups using spatial overlap and feature similarity, then fuse them with a learned merger~$\mathcal{M}^\text{mrg}$.

For each Feature Gaussian~$\tilde{\mathcal{G}}_j$, we retrieve the $k$ nearest neighbors via 3D $k$NN on positions, excluding candidates from the same input view.
A candidate pair passes the match gate if both
(i)~the cosine similarity between latent features exceeds~$\tau_f$, and
(ii)~the AABB intersection-over-union of the FGs' geometry exceeds~$\tau_g$.
This produces a set of matched edges over all Feature Gaussians.

Since one FG may match several others, and matching is not transitive, we take the connected components of this edge graph as the final merge groups.
Feature Gaussians with no matches are appended to the final set $\tilde{\mathcal{G}}$ unchanged.
$\mathcal{M}^\text{mrg}$ merges each connected group using the same SAB$+$PMA~\cite{Lee2019} backbone as the encoder (\figref{fig:architecture}).
Dedicated zero-initialized residual heads predict parameter updates to the $0^\text{th}$ Feature Gaussian in the group: $\Delta\boldsymbol{\mu}, \Delta\mathbf{z}, \Delta m, \Delta\mathbf{c}_0, \Delta\mathbf{s}, \Delta\mathbf{q}$.
This way the merger starts as an identity and training remains stable.

\subsection{Refinement and Online Reconstruction}
\label{sec:refinement_online}

\textbf{Refinement.}
The final rendering quality depends in part on how well neighboring Feature Gaussians~$\tilde{\mathcal{G}}$ interact — particularly along thin structures or at depth boundaries.
In these areas, individually compressed groups might not perfectly line up in the final reconstruction. 
The refiner $\mathcal{M}^\text{ref}$ addresses this by updating each Feature Gaussian~$\tilde{\mathcal{G}}_j$ based on its $k$NN neighborhood through self-attention: neighbors' relative positions and attributes serve as context tokens, and residual heads predict parameter updates $\Delta\boldsymbol{\mu}, \Delta\mathbf{z}, \Delta m, \Delta\mathbf{c}_0, \Delta\mathbf{s}, \Delta\mathbf{q}$.
As with the merger, all residual projections are zero-initialized for stable training.

\textbf{Online reconstruction.}
Our pipeline also supports online reconstruction, where views are processed as a sequence as they arrive rather than in a single batch.
This allows for updates to the reconstructed scene as new observations become available and enables reconstruction of large scenes that do not fit into memory at once.
In practice, we initialize the Feature Gaussian set $\tilde{\mathcal{G}}$ from a first batch of observations.
When a new view arrives, the pipeline groups and encodes its per-pixel Gaussians into new Feature Gaussians, then matches and merges them into the existing set $\tilde{\mathcal{G}}$.
The decoder produces output Gaussians on demand from the current state of $\tilde{\mathcal{G}}$ via $\mathcal{M}^\text{dec}_K$.

\subsection{Training Objectives}
\label{sec:training}

During training, the feed-forward backbone $\mathcal{M}^\text{FF}$ remains frozen; we optimize only the encoder $\mathcal{M}^\text{enc}$, merger $\mathcal{M}^\text{mrg}$, refiner $\mathcal{M}^\text{ref}$, and decoders $\mathcal{M}^\text{dec}_K$.
We train the pipeline end-to-end on multi-view image data with the following losses.

\textbf{Photometric reconstruction.}
The primary supervision consists of Mean Squared Error (MSE), SSIM~\cite{Wang2004}, and LPIPS~\cite{Zhang2018} losses between rendered and ground-truth images, optionally computed on both input views and held-out novel views.

\textbf{Teacher loss.}
Due to randomly initialized weights of our modules, the geometry of the compressed Gaussians is in turn essentially random during early training.
This can cause single Gaussians to cover the renders from the training views entirely, inhibiting gradient flow. 
To address this, we supervise the encoder with a closed-form moment-matching target~\cite{kerbl2024hierarchical}: for each superpixel group $\mathcal{S}_j$, we fit a single Gaussian to the group's opacity-weighted mean
\begin{equation}
\label{eq:teacher_mean}
\boldsymbol{\mu}^\text{teach}_j = \frac{\sum_{i \in \mathcal{S}_j} \alpha_i \, \boldsymbol{\mu}_i}{\sum_{i \in \mathcal{S}_j} \alpha_i}
\end{equation}
and opacity-weighted covariance
\begin{equation}
\label{eq:teacher_cov}
\boldsymbol{\Sigma}^\text{teach}_j = \frac{\sum_{i \in \mathcal{S}_j} \alpha_i \, (\boldsymbol{\mu}_i - \boldsymbol{\mu}^\text{teach}_j)(\boldsymbol{\mu}_i - \boldsymbol{\mu}^\text{teach}_j)^\top}{\sum_{i \in \mathcal{S}_j} \alpha_i}.
\end{equation}
We compare the resulting Gaussian parameters to the output of our merging pipeline using a weighted sum of a geodesic quaternion loss for $\mathbf{q}_i$ and L1 losses for the remaining parameters. This teacher signal decays on a schedule, allowing the learned encoder to eventually surpass the moment-matched initialization.

\textbf{Decoder diversification.}
For $K\!>\!1$ decoders, a regularizer encourages the $K$ output Gaussians to be spatially diverse without collapsing into identical copies.
We define a target AABB-IoU and apply an attraction/repulsion term around it, and additionally penalize large opacity differences within a group to prevent single-primitive dominance.
Decaying this loss during training lets the photometric objective override it once diversity is established, allowing the model to collapse to fewer Gaussians where a single primitive suffices.
The full formulation of this regularizer is given in the supplementary material (\secref{sec:diversification}).

The complete training objective combines all terms:
\begin{equation}
\label{eq:total_loss}
\mathcal{L} = \lambda_\text{MSE}\mathcal{L}_\text{MSE} + \lambda_\text{SSIM}\mathcal{L}_\text{SSIM} + \lambda_\text{LPIPS}\mathcal{L}_\text{LPIPS} + \lambda_\text{teach}\mathcal{L}_\text{teach} + \lambda_\text{div}\mathcal{L}_\text{div},
\end{equation}
where $\lambda_\text{teach}$ and $\lambda_\text{div}$ follow a decay schedule and concrete weight values are provided in the supplementary material (\secref{sec:hyperparams}).

\section{Experimental Evaluation}
\label{sec:exp}

We evaluate our saliency-guided superpixel-based primitive merging strategy on novel view synthesis quality and number of primitives.

\subsection{Experimental Setup}
\label{sec:exp_setup}

\textbf{Implementation Details.}
We evaluate our approach with three complementary FF backbones: DepthSplat~\cite{Xu2024c}, a posed method that leverages multi-view cost volumes augmented with monocular depth features; AnySplat~\cite{Jiang2025}, an unposed method that jointly estimates geometry from unconstrained views; and Depth Anything~3 (DA3)~\cite{Lin2025a}, a foundation model that unifies depth estimation, pose estimation, and per-pixel 3D Gaussian prediction in a single ViT backbone, supporting both posed and unposed operation.
We train all added modules with the AdamW optimizer~\cite{Loshchilov2017DecoupledWD} for 300\,000 steps on the DL3DV-10K dataset~\cite{Ling2024}, using a cosine schedule with a maximum learning rate of $5\times10^{-5}$.
For the level-of-detail decoder, we maintain three heads with $K\!=\!1$, $K\!=\!2$, and $K\!=\!4$ output Gaussians per Feature Gaussian, which are jointly trained.
We keep the backbones frozen during training.

\textbf{Evaluation Details.}
We test the reconstruction quality on DL3DV-Bench~\cite{Ling2024}, which contains 140 scenes not used during training.
We additionally evaluate on MipNeRF360~\cite{Barron2022} and Tanks and Temples~\cite{Knapitsch2017}, two widely used benchmarks that were not included in the training data.
We sample input views at regular intervals along the camera trajectory, with novel views placed in-between.
Images are resized so that the longer side is 476 pixels.
We mask out areas of the scene not visible in any input view and exclude them from metric computation to focus evaluation on the quality of the reconstruction rather than the extrapolation.
Each experiment uses 3, 6, 9, and 12 input views; results are averaged across all scenes and view splits.
Detailed per-view-count results (\secref{sec:per_view}) and input view reconstruction metrics (\secref{sec:recon_tables}) appear in the supplementary material, which also reports results computed without visibility masking (\secref{sec:eval_protocol}).
We report PSNR, SSIM~\cite{Wang2004}, and LPIPS~\cite{Zhang2018} for novel view synthesis.
For efficiency we report the relative primitive count $r_c$ (fraction of the unmerged count that is retained), which directly correlates with rendering time and memory usage.
Further evaluation details are provided in the supplementary material (\secref{sec:eval_protocol}).

In addition to reporting the quality of the reconstruction directly produced by the three backbones, we compare against ReSplat~\cite{Xu2025a}, which predicts Gaussians in a $16\times$ subsampled space and refines them recurrently, and VolSplat~\cite{wang2025volsplat}, a voxel-based method producing Gaussian in a regular grid.
We additionally report the performance of AnySplat's built-in post-hoc voxelization and of heuristic moment matching of Gaussians grouped into superpixels.
Due to unavailable public code, we could not include Off The Grid~\cite{Moreau2025} and Fuse-and-Refine~\cite{wang2025learning} in our evaluation.

\subsection{Main Results}
\label{sec:exp_results}

\begin{table}[t]
\centering
\caption{Novel view synthesis quality for all backbone variants and baselines.
Best in \textbf{bold}, second best \underline{underlined}.}
\vspace{0.2cm}
\label{tab:nvs}
\setlength{\tabcolsep}{2.2pt}
\scriptsize
\begin{tabular}{@{}ll r ccc ccc ccc@{}}
\toprule
& & & \multicolumn{3}{c}{DL3DV-Bench} & \multicolumn{3}{c}{MipNeRF360} & \multicolumn{3}{c}{T\&T} \\
\cmidrule(lr){4-6}\cmidrule(lr){7-9}\cmidrule(lr){10-12}
Backbone & Method & $r_c$\,(\%)$\downarrow$ & PSNR$\uparrow$ & SSIM$\uparrow$ & LPIPS$\downarrow$ & PSNR$\uparrow$ & SSIM$\uparrow$ & LPIPS$\downarrow$ & PSNR$\uparrow$ & SSIM$\uparrow$ & LPIPS$\downarrow$ \\
\midrule
\multirow{6}{*}{AnySplat} & Unmerged             &    100.0\% &                 13.73 &                 0.361 &                 0.495 &                 12.17 &                 0.345 &                 0.497 &                 11.89 &                 0.416 &                 0.466 \\
 & Voxelized            &     83.2\% &                 13.76 &                 0.360 &                 0.497 &                 12.17 &                 0.344 &                 0.498 &                 11.89 &                 0.413 &                 0.468 \\
\cmidrule{2-12}
 & Mom.\ Match.         &      5.3\% &                 14.45 &                 0.410 &                 0.566 &                 13.46 &                 0.398 &                 0.559 &                 12.49 &                 0.463 &                 0.517 \\
 & Ours ($K\!=\!1$)     & \underline{4.4\%} &                 14.70 &                 0.408 &                 0.550 &                 14.48 &                 0.405 &                 0.544 &                 13.24 &                 0.458 &                 0.504 \\
 & Ours ($K\!=\!2$)     &      8.9\% &                 14.81 &                 0.411 &                 0.549 &                 14.74 &                 0.406 &                 0.544 &                 13.81 &                 0.473 &                 0.505 \\
 & Ours ($K\!=\!4$)     &     17.8\% &                 14.78 &                 0.409 &                 0.547 &                 14.81 &                 0.408 &                 0.542 &                 13.93 &                 0.474 &                 0.503 \\
\midrule
\multirow{5}{*}{DepthSplat} & Unmerged             &    100.0\% &                 15.78 &                 0.577 &     \underline{0.347} &                 13.96 &                 0.480 &     \underline{0.389} &                 13.72 &                 0.549 &     \underline{0.377} \\
\cmidrule{2-12}
 & Mom.\ Match.         &      6.1\% &                 13.16 &                 0.466 &                 0.573 &                 12.35 &                 0.424 &                 0.569 &                 11.43 &                 0.498 &                 0.550 \\
 & Ours ($K\!=\!1$)     &      5.3\% &                 17.17 &                 0.529 &                 0.440 &                 15.37 &                 0.457 &                 0.452 &                 14.33 &                 0.508 &                 0.461 \\
 & Ours ($K\!=\!2$)     &     10.5\% &                 17.41 &                 0.536 &                 0.436 &                 16.05 &                 0.469 &                 0.448 &                 14.60 &                 0.517 &                 0.458 \\
 & Ours ($K\!=\!4$)     &     21.1\% &                 17.50 &                 0.539 &                 0.433 &                 16.31 &                 0.474 &                 0.446 &                 14.78 &                 0.522 &                 0.457 \\
\midrule
\multirow{5}{*}{DA3} & Unmerged             &    100.0\% &        \textbf{18.68} &        \textbf{0.620} &        \textbf{0.297} &                 16.71 &                 0.495 &        \textbf{0.356} &                 15.07 &                 0.546 &        \textbf{0.353} \\
\cmidrule{2-12}
 & Mom.\ Match.         &      4.9\% &                 13.79 &                 0.459 &                 0.637 &                 14.04 &                 0.440 &                 0.601 &                 12.55 &                 0.505 &                 0.611 \\
 & Ours ($K\!=\!1$)     & \textbf{4.4\%} &                 18.37 &                 0.578 &                 0.422 &                 17.87 &                 0.514 &                 0.442 &                 16.00 &                 0.562 &                 0.421 \\
 & Ours ($K\!=\!2$)     &      8.8\% &                 18.50 &                 0.582 &                 0.417 &     \underline{18.02} &     \underline{0.517} &                 0.440 &     \underline{16.09} &     \underline{0.566} &                 0.418 \\
 & Ours ($K\!=\!4$)     &     17.3\% &     \underline{18.54} &     \underline{0.583} &                 0.416 &        \textbf{18.06} &        \textbf{0.518} &                 0.439 &        \textbf{16.09} &        \textbf{0.566} &                 0.418 \\
\midrule
\midrule
\multirow{2}{*}{ReSplat} & Init                 &      6.2\% &                 13.17 &                 0.380 &                 0.556 &                 15.57 &                 0.369 &                 0.566 &                 12.37 &                 0.383 &                 0.591 \\
 & Recurrent~4          &      6.2\% &                 15.39 &                 0.468 &                 0.519 &                 16.79 &                 0.418 &                 0.558 &                 13.38 &                 0.437 &                 0.578 \\
VolSplat & ---                  &     93.4\% &                 14.12 &                 0.366 &                 0.607 &                 11.81 &                 0.251 &                 0.700 &                 11.71 &                 0.320 &                 0.650 \\
\bottomrule

\end{tabular}
\vspace{-1cm}
\end{table}

\textbf{Novel View Synthesis Quality.}
\tabref{tab:nvs} reports novel view synthesis results for all methods.
Using our pipeline with the strongest backbone (DA3), the $K\!=\!1$ decoder achieves \psnrdaThreemergedkone\,dB PSNR averaged over all benchmarks at $r_c = \rcdaThreemergedkone$\%, even slightly surpassing the DA3 unmerged quality at \psnrdaThreeunmerged\,dB.
The gain in performance over DA3 is only achieved on MipNeRF360 and Tanks~\&~Temples, which have much larger viewpoint changes than DL3DV-Bench.
This reflects the regularising effect of superpixel-based merging: fusing per-pixel Gaussians into compact, spatially coherent groups can remove floaters and overly large Gaussians, and reduces high-frequency noise.
These effects are consistent across all backbones, with the largest gain for DepthSplat, which sometimes produces overly large Gaussians that strongly impact the reconstruction quality.
LPIPS responds very sensitively to texture degradation from merging, and therefore our pipeline does not fully preserve it.
ReSplat achieves a comparable $r_c$ of \rcresplatresplat\% but with lower quality than our method, and without the flexibility of adapting to more powerful foundation models, as they become available.
VolSplat performs significantly worse than the other methods, and fails to reproduce the reconstruction quality reported in its original paper.
This is likely due to the fact that VolSplat was exclusively trained on RealEstate-10K~\cite{Zhou2018}, which mostly contains scenes with smaller viewpoint changes, different to our benchmarks.

\textbf{Primitive Count vs.\ Quality Trade-off.}
\figref{fig:pareto} shows the quality--efficiency trade-off for the DA3 backbone, averaged over all three benchmarks.
Our method ($K\!=\!1,2,4$) traces a favorable curve that stays close to the unmerged quality ceiling with $r_c$ between \rcdaThreemergedkone\% and \rcdaThreemergedkfour\%.
ReSplat and VolSplat produce suboptimal results, failing to achieve the same quality as our method at comparable or higher primitive counts.

\begin{figure*}[t]
\centering
\newlength{\qualimw}
\setlength{\qualimw}{0.192\textwidth}
\setlength{\tabcolsep}{1pt}
\begin{tabular}{@{}ccccc@{}}
\toprule
\input{pics/gen/qualitative_nvs}
\end{tabular}
\caption{Qualitative novel view synthesis comparison at 12 input views (DA3 backbone).
Unseen areas are highlighted in red and excluded from metric computation.}
\label{fig:qualitative}
\vspace{-0.5cm}
\end{figure*}

\textbf{Qualitative Comparisons.}
\figref{fig:qualitative} shows qualitative novel view synthesis results at 12 input views across all three benchmarks.
Our method ($K\!=\!1$, DA3 backbone) largely reproduces the unmerged quality at a fraction of the primitive count.
ReSplat occasionally yields sharper detail but introduces more severe structured artifacts elsewhere.
VolSplat fully fails to reconstruct the scene structure for many scenes.

\begin{figure}[t]
\centering
\input{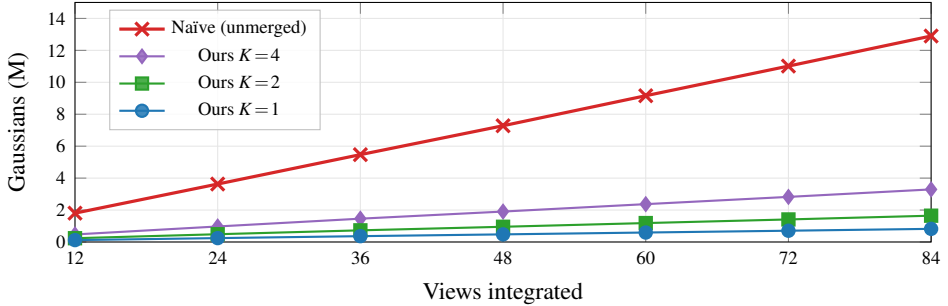}%
\definecolor{onlineColKone}{HTML}{1F77B4}    %
\definecolor{onlineColKtwo}{HTML}{2CA02C}    %
\definecolor{onlineColKfour}{HTML}{9467BD}   %
\definecolor{onlineColNaive}{HTML}{D62728}   %

\begin{tikzpicture}
\begin{axis}[
  width=\linewidth,
  height=4.75cm,
  xlabel={\small Views integrated},
  ylabel={\small Gaussians (M)},
  xmin=\onlineXmin, xmax=\onlineXmax,
  ymin=0, ymax=\onlineYmax,
  xtick={12,24,36,48,60,72,84},
  ytick={0,2,4,6,8,10,12,14},
  grid=major,
  grid style={gray!20},
  tick label style={font=\scriptsize},
  label style={font=\small},
  legend style={
    font=\scriptsize,
    at={(0.04,0.97)},
    anchor=north west,
    draw=gray!50,
    fill=white,
    fill opacity=0.85,
    text opacity=1,
    row sep=0pt,
  },
  clip=false,
]

\addplot[
  color=onlineColNaive,
  mark=x,
  mark size=3.5pt,
  very thick,
] coordinates {\onlineNaiveCoords};
\addlegendentry{Naïve (unmerged)}

\addplot[
  color=onlineColKfour,
  mark=diamond*,
  mark size=2.5pt,
  thick,
] coordinates {\onlineKfourCoords};
\addlegendentry{Ours $K\!=\!4$}

\addplot[
  color=onlineColKtwo,
  mark=square*,
  mark size=2.3pt,
  thick,
] coordinates {\onlineKtwoCoords};
\addlegendentry{Ours $K\!=\!2$}

\addplot[
  color=onlineColKone,
  mark=*,
  mark size=2.3pt,
  thick,
] coordinates {\onlineKoneCoords};
\addlegendentry{Ours $K\!=\!1$}

\end{axis}
\end{tikzpicture}

\vspace{-0.15cm}
\caption{Gaussian count vs.\ integrated views in the online setting, averaged over 7 MipNeRF360 scenes.}
\label{fig:online}
\end{figure}

\begin{table}[t]
\centering
\caption{Online reconstruction quality at end-of-sequence, averaged over all 9 MipNeRF360 scenes (target views).
Best per column in \textbf{bold}.}
\vspace{0.2cm}
\label{tab:online}
\setlength{\tabcolsep}{4pt}
\scriptsize
\begin{tabular}{@{}l r r ccc r@{}}
\toprule
Method & \#Prim.$\downarrow$ & $r_c$\,(\%)$\downarrow$ & PSNR$\uparrow$ & SSIM$\uparrow$ & LPIPS$\downarrow$ & FPS$\uparrow$ \\
\midrule
Na\"ive backbone & 16.1\,M & 100.0 & 15.27 & \textbf{0.372} & \textbf{0.560} & 60 \\
\midrule
Ours $K\!=\!1$ & \textbf{1.0\,M} & \textbf{6.2} & 15.33 & 0.367 & 0.612 & \underline{365} \\
Ours $K\!=\!2$ & \underline{2.0\,M} & \underline{12.4} & \underline{15.41} & 0.368 & 0.611 & \textbf{378} \\
Ours $K\!=\!4$ & 4.0\,M & 24.8 & \textbf{15.45} & \underline{0.370} & \underline{0.610} & 245 \\
\bottomrule

\end{tabular}
\vspace{-0.5cm}
\end{table}

\textbf{Online Reconstruction.}
We evaluate our pipeline in an online setting where views are integrated incrementally.
Starting from an empty scene, batches of 12 views are processed in sequence, in the order of the camera trajectory, until all context views are integrated.
Half of the total views are used as input and the rest as target views for evaluation.
We report primitive growth averaged over the 7 scenes from MipNeRF360 that provide at least 84 context views (bicycle, bonsai, counter, flowers, garden, kitchen, room), and novel view synthesis quality at the end of the sequence averaged over all 9 MipNeRF360 scenes.
\figref{fig:online} compares the cumulative Gaussian count of our three decoders against the naïve baseline, which accumulates all raw per-pixel Gaussians without merging.
The naïve baseline grows at $\sim$\onlineNaivePerStepM\,M Gaussians per 12 views; our $K\!=\!1$ decoder grows at $\sim$\onlineKonePerStepM\,M per step — a $\onlineSlopeRatioX\times$ reduction in slope owing to within-step compression from superpixel grouping, with additional savings from cross-step merging of overlapping regions.
At 84 integrated views, $K\!=\!1$ retains \onlineFinalKoneM\,M Gaussians against \onlineFinalNaiveM\,M for the naïve baseline.
\tabref{tab:online} summarises per-scene-average quality at the end of each sequence.
Despite holding \onlinePrimRatioX$\times$ fewer primitives, our $K\!=\!1$ decoder closely approaches the naïve backbone on PSNR and SSIM while trading off some perceptual fidelity (LPIPS), and renders \onlineFpsSpeedupX$\times$ faster.

\subsection{Ablation Studies}
\label{sec:ablations}

Replacing our learned merger with heuristic moment matching to combine the Gaussians within each superpixel performs worse across all metrics, confirming that a learned representation is essential for quality-preserving fusion.
The magnitude of the benefit of the learned merging pipeline depends on the uniformity of the Gaussians predicted by the backbone: with the more uniform AnySplat Gaussians, moment matching performs better and the gap to our method is smaller (\tabref{tab:nvs}).

All further ablations use the AnySplat backbone and report the $K\!=\!1$ decoder averaged over DL3DV-Bench, MipNeRF360, and Tanks~\&~Temples.
\tabref{tab:ablation} summarizes all variants against the unmerged reference.
Removing the refiner $\mathcal{M}^\text{ref}$ degrades all metrics (\tabref{tab:ablation}), indicating that the refiner contributes significantly to overall performance.
Disabling cross-view merging slightly improves PSNR at the cost of a higher primitive count, suggesting that cross-view fusion reduces redundancy at the cost of a slight quality drop.
Within-view merging is the dominant source of compression in our pipeline, while cross-view merging is an optional addition that provides a smaller, but not negligible, further reduction: relative to within-view merging alone, cross-view merging retains \rcCrossViewRetained\% of the Gaussians, a compression roughly comparable to AnySplat's built-in post-hoc voxelization (\rcanysplatVarFeatvoxelized\%, \tabref{tab:nvs}).

\begin{table}[t]
\centering
\caption{Ablation study (AnySplat backbone, $K\!=\!1$, averaged over all benchmarks).
Best in \textbf{bold}, second best \underline{underlined}.}
\vspace{0.2cm}
\label{tab:ablation}
\setlength{\tabcolsep}{4pt}
\scriptsize
\begin{tabular}{@{}l r ccc@{}}
\toprule
Variant & $r_c$\,(\%)$\downarrow$ & PSNR$\uparrow$ & SSIM$\uparrow$ & LPIPS$\downarrow$ \\
\midrule
\textit{Backbone (unm.)} &    100.0\% &                   12.59 &                   0.374 &                   0.486 \\
\midrule
Ours (full)               & \underline{4.4\%} &          \textbf{14.14} &                   0.423 &                   0.533 \\
w/o refiner               &      4.7\% &                   13.49 &                   0.389 &                   0.534 \\
w/o cross-view            &      5.3\% &       \underline{14.13} &       \underline{0.424} &       \underline{0.531} \\
zero shot larger SPs      & \textbf{2.0\%} &                   13.97 &          \textbf{0.424} &                   0.548 \\
zero shot smaller SPs     &      8.7\% &                   14.11 &                   0.420 &          \textbf{0.523} \\
\bottomrule

\end{tabular}
\vspace{-0.5cm}
\end{table}

\begin{figure}[t]
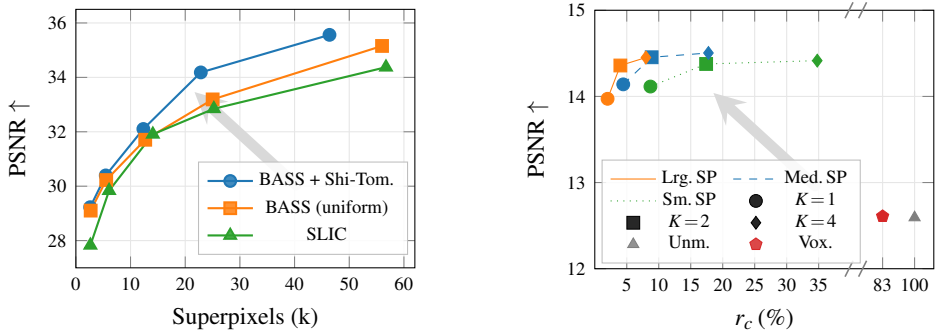

\centering
\begin{minipage}[b]{0.47\linewidth}
  \centering
  \input{pics/gen/sp_proxy_pareto_data.tex}%
  \definecolor{spColBassShitom}{HTML}{1F77B4}    %
\definecolor{spColBassNoShitom}{HTML}{FF7F0E}  %
\definecolor{spColSlic}{HTML}{2CA02C}          %

\begin{tikzpicture}
\begin{axis}[
  width=\linewidth,
  height=5.0cm,
  xlabel={\small Superpixels (k)},
  ylabel={\small PSNR $\uparrow$},
  xmin=0, xmax=62,
  ymin=27.0, ymax=36.5,
  xtick={0,10,20,30,40,50,60},
  xticklabels={0,10,20,30,40,50,60},
  ytick={28,30,32,34,36},
  grid=major,
  grid style={gray!20},
  tick label style={font=\scriptsize},
  label style={font=\small},
  legend style={
    font=\scriptsize,
    at={(0.98,0.05)},
    anchor=south east,
    draw=gray!50,
    fill=white,
    fill opacity=0.85,
    text opacity=1,
    row sep=0pt,
  },
  clip=false,
]
\draw[->, >=stealth, gray!30, line width=4pt, line cap=round]
  (rel axis cs:0.65, 0.32) -- (rel axis cs:0.35, 0.68);  

\addplot[
  color=spColBassShitom,
  mark=*,
  mark size=2.2pt,
  thick,
] coordinates {\spProxyBASSShiTomCoords};
\addlegendentry{BASS + Shi-Tom.}

\addplot[
  color=spColBassNoShitom,
  mark=square*,
  mark size=2.2pt,
  thick,
] coordinates {\spProxyBASSuniformCoords};
\addlegendentry{BASS (uniform)}

\addplot[
  color=spColSlic,
  mark=triangle*,
  mark size=2.5pt,
  thick,
] coordinates {\spProxySLICCoords};
\addlegendentry{SLIC}

\end{axis}
\end{tikzpicture}
\end{minipage}\hfill
\begin{minipage}[b]{0.47\linewidth}
  \centering
  \input{pics/gen/sp_size_pareto_data.tex}%
  \input{pics/segmentation/sp_size_pareto.tex}
\end{minipage}
\caption{%
  \textit{Left:} Proxy PSNR vs.\ superpixel count for three segmentation algorithms (100 scenes~$\times$~10 views).
  Each pixel is replaced by its superpixel mean color and compared to the original image.
  \textit{Right:} NVS PSNR vs.\ relative primitive count for three SP sizes (AnySplat backbone, zero-shot, averaged over all benchmarks). Arrows indicate favorable direction.}
\label{fig:sp_proxy_pareto}
\vspace{-0.25cm}
\end{figure}

\textbf{Superpixel Algorithm Comparison.}
To evaluate the segmentation quality of different superpixel algorithms across a wide range of segment counts, without the cost of the full pipeline evaluation, we use a proxy metric: each pixel in an image is replaced by the mean color of its superpixel, and the resulting image is compared to the ground truth.
This measures how faithfully a given superpixel layout can represent the visual image content, which determines the upper bound on merging quality.
We apply this proxy to 1\,000 images (100 scenes~$\times$~10 views) from the training set and report PSNR against the original pixels.
\figref{fig:sp_proxy_pareto} shows the quality vs.\ superpixel count trade-off for BASS~\cite{uziel2019bayesian} with uniform seeding, BASS with our Shi-Tomasi corner detector seeding, and SLIC~\cite{Achanta2010}.
BASS with Shi-Tomasi corner seeding consistently dominates the other algorithms at all budgets, achieving higher proxy PSNR at equal superpixel count.
Based on these results, we evaluate the full pipeline (without retraining) with BASS and Shi-Tomasi seeding at three superpixel size settings, which yield ${\sim}$\spCountSmall, ${\sim}$\spCountMedium, and ${\sim}$\spCountLarge\ superpixels per image on average.

Larger superpixels reduce the relative primitive count $r_c$ at the cost of quality, \eg, from \psnranysplatVarFeatmergedkone\,dB PSNR to \psnrablfeatvariationalsplargemergedkone\,dB PSNR at roughly half the $r_c$ (\tabref{tab:ablation}).
This confirms that finer superpixels better preserve reconstruction quality, at the cost of a higher $r_c$, with diminishing returns at very small superpixel sizes.
Using larger superpixels therefore provides a similar trade-off to using the higher $K$ decoder heads, with the decoder heads being slightly more effective at the same $r_c$.
For example, the $K\!=\!2$ decoder achieves \psnranysplatVarFeatmergedktwo\,dB PSNR at $r_c=\rcanysplatVarFeatmergedktwo$\%, while the $K\!=\!4$ decoder with larger superpixels achieves \psnrablfeatvariationalsplargemergedkfour\,dB PSNR at a similar $r_c=\rcablfeatvariationalsplargemergedkfour$\%.
This provides the user with two complementary mechanisms to flexibly adjust the quality--efficiency trade-off according to their needs, both without retraining.

\section{Conclusion}
\label{sec:conclusion}

We present a structure-aware approach for strongly compressing Gaussian splatting representations generated by feed-forward networks. 
Our superpixel-based primitive merging strategy intelligently consolidates per-pixel Gaussians into a compact, content-adaptive representation.
We group Gaussians via saliency-guided superpixel segmentation, encode each group into a Feature Gaussian with a learned encoder, and match and merge Feature Gaussians across views.
A level-of-detail decoder provides explicit control over the output primitive count at inference, enabling flexible quality-efficiency trade-offs.
The pipeline operates as a backbone-agnostic post-processing module that retains the bulk of the reconstruction quality across three benchmarks and three feed-forward backbones at just $\nicefrac{1}{20}^\text{th}$ of the Gaussians, accepting a slight reduction in fine-detail fidelity in exchange for dramatically improved efficiency.
Compared to other reduced-primitive methods, our approach provides more robust and higher quality reconstructions, particularly in sparse-view settings.
Additionally, for large viewpoint changes, the merged primitives can even improve some metrics by suppressing common FF artifacts such as semi-transparent fog, or high-frequency noise.

\textbf{Limitations.}
Our method depends on the quality of the underlying feed-forward reconstruction; if the initial primitives are of low quality, merging may not recover a good representation.
The matching and merging pipeline introduces a computational overhead of $\timingMergingOverheaddaThree$\,ms compared to direct feed-forward inference (details in the supplementary material, \secref{sec:timing}).
While SSIM and PSNR metrics are preserved or improved, the LPIPS metric is slightly degraded due to some loss of fine detail in the merged representation.

\textbf{Future work.}
Promising directions include extensions to dynamic scenes where temporal consistency must be maintained, and replacing the algorithmic superpixel grouping with a learned grouping mechanism that can be optimized jointly with the merging pipeline.
LPIPS degradation could potentially be mitigated by using a richer output representation, such as Textured Gaussians~\cite{Chao2025} or 3D Convex Splatting~\cite{Held2025}.

\textbf{Acknowledgements.}
This work has partially been funded by the Deutsche Forschungsgemeinschaft (DFG, German Research Foundation) under Germany's Excellence Strategy, EXC-2070 -- 390732324 -- PhenoRob and by the German Federal Ministry of Research, Technology and Space~(BMFTR) under the Robotics Institute Germany~(RIG).
As part of the IPCEI ME/CT the project is supported by the Federal Ministry for Economic Affairs and Energy, by the Ministry for Economic Affairs, Labor and Tourism of Baden-Württemberg and financed by the European Union -- NextGenerationEU.

\bibliography{main}

\appendix
\clearpage
\setcounter{page}{1}
\maketitlesupplementary

\renewcommand{\thefigure}{S\arabic{figure}}
\renewcommand{\thetable}{S\arabic{table}}
\renewcommand{\thesection}{\Alph{section}}

In this supplementary material we provide additional details that expand on the main paper.
\secref{sec:eval_protocol} gives the full evaluation protocol and reports results without visibility masking.
\secref{sec:hyperparams} lists the training hyperparameters including loss weights and decay schedules.
\secref{sec:timing} provides a timing analysis of the pipeline components, together with storage size and the rendering-vs-reconstruction break-even point.
\secref{sec:diversification} describes the decoder diversification regularizer.
\secref{sec:encoder_compute} explains how the encoder's compute budget is bounded for large superpixels.
\secref{sec:recon_tables} provides input-view reconstruction metrics.
\secref{sec:per_view} provides a per-view-count breakdown of novel view synthesis quality.
\secref{sec:extensions} discusses extensions to the method, including content-adaptive per-superpixel $K$ and compaction of voxel-based FF backbones.
\secref{sec:qualitative_suppl} provides additional qualitative results.
\secref{sec:da3_posed_unposed} compares the posed and unposed DA3 backbone configurations.
We have also included the code for training and evaluation of our method. Please refer to the README for information on setup and usage. We will make the code public upon acceptance, after internal review.

\section{Evaluation Protocol Details}
\label{sec:eval_protocol}

We provide the full details of our evaluation protocol here to allow independent reproduction.

\textbf{Image resolution.}
All images (input and ground-truth target views) are resized so that the longer side equals 476\,px, preserving the original aspect ratio without any cropping.

\textbf{Visibility masking.}
None of our three benchmarks provide ground-truth scene geometry, so we cannot directly determine which regions of a target view are unobserved by the input views.
Instead, we estimate a depth map for every input view using Depth Anything~3~\cite{Lin2025a}, reproject each input view's depth into every target view using the known (or estimated, for unposed backbones) camera parameters, and mark a target pixel as visible if it receives at least one reprojected sample.
To account for depth estimation uncertainty near depth discontinuities, we dilate the resulting binary visibility mask with a $3\times3$ max-pooling kernel before excluding unmarked pixels from all metric computations.

\textbf{View sampling.}
Context (input) views are sampled at equal intervals along the full camera trajectory of each scene, for each of the four view counts (3, 6, 9, 12) evaluated in the main paper.
Target (novel) views are placed at the midpoint between each pair of consecutive context views.
Because scenes have varying trajectory lengths, the exact frame indices used differ from scene to scene, but the sampling rule itself is deterministic and reproducible given the scene's frame count.

\textbf{Results without visibility masking.}
\tabref{tab:unmasked} reports PSNR, SSIM, and LPIPS without visibility masking (\ie, computed over the full target image, including regions unobserved by any input view) for the DA3 backbone, averaged over all three benchmarks and all four input view counts (3, 6, 9, 12), matching the aggregation used throughout the main evaluation.
The relative ordering of methods is consistent with the masked results in \tabref{tab:nvs}, confirming that our masking protocol does not unfairly favor any particular method.

\begin{table}[t]
\centering
\caption{Novel view synthesis quality without visibility masking (DA3 backbone, averaged over all three benchmarks and all four input view counts).
Best in \textbf{bold}, second best \underline{underlined}.
Compare to the masked results in \tabref{tab:nvs}.}
\vspace{0.2cm}
\label{tab:unmasked}
\scriptsize
\begin{tabular}{@{}l ccc@{}}
\toprule
Method & PSNR$\uparrow$ & SSIM$\uparrow$ & LPIPS$\downarrow$ \\
\midrule
Ours ($K\!=\!4$) & \textbf{\psnrdaThreemergedkfourUnmasked} & \textbf{\ssimdaThreemergedkfourUnmasked} & \underline{\lpipsdaThreemergedkfourUnmasked} \\
Ours ($K\!=\!1$) & \underline{\psnrdaThreemergedkoneUnmasked} & \underline{\ssimdaThreemergedkoneUnmasked} & \lpipsdaThreemergedkoneUnmasked \\
Mom.\ Match. & \psnrdaThreemomentmatchingUnmasked & \ssimdaThreemomentmatchingUnmasked & \lpipsdaThreemomentmatchingUnmasked \\
DA3 (unmerged) & \psnrdaThreeunmergedUnmasked & \ssimdaThreeunmergedUnmasked & \textbf{\lpipsdaThreeunmergedUnmasked} \\
ReSplat & \psnrresplatresplatUnmasked & \ssimresplatresplatUnmasked & \lpipsresplatresplatUnmasked \\
VolSplat & \psnrvolsplatvolsplatUnmasked & \ssimvolsplatvolsplatUnmasked & \lpipsvolsplatvolsplatUnmasked \\
\bottomrule
\end{tabular}
\end{table}

\section{Training Hyperparameters}
\label{sec:hyperparams}

\tabref{tab:hyperparams} summarizes the loss weights and their decay schedules.
We train all added modules jointly for 300\,000 steps with AdamW~\cite{Loshchilov2017DecoupledWD} on DL3DV-10K~\cite{Ling2024} using a randomly sampled subset of 1 to 12 input views.
For depthsplat we use 2 to 12 views as the backbone does not support single view input.
The learning rate follows a cosine schedule from a peak of $5\times10^{-5}$ after 2\,000 linear warmup steps down to a minimum of $10^{-6}$.
We apply gradient clipping with a maximum norm of $10.0$.

\section{Timing, Storage, and Rendering Throughput}
\label{sec:timing}

We measured runtimes on a single NVIDIA B200 GPU at 12 input views at $476 \times 476$ resolution, averaged over DL3DV-Bench, MipNeRF360, and T\&T.
They cover the full inference path from input images to output Gaussians, excluding data loading and metric computation.

The BASS superpixel algorithm with Shi-Tomasi seeding runs at $\bassTimingMs$\,ms per image.
For a 12-view scene this amounts to approximately $\bassSegOverheadTwelveViews$\,ms of segmentation overhead, independent of backbone choice.
The backbone inference time spans from $\timingBackboneanysplatVarFeatunmerged$\,ms (AnySplat) to $\timingBackbonedepthsplatunmerged$\,ms (DepthSplat), with the DA3 backbone in between at $\timingBackbonedaThreeunmerged$\,ms.
Including superpixel segmentation and backbone inference, the total reconstruction time for deployment of our pipeline is approximately $\timingTotalDeployanysplatVarFeat$\,ms (AnySplat backbone), $\timingTotalDeploydepthsplat$\,ms (DepthSplat backbone), and $\timingTotalDeploydaThree$\,ms (DA3 backbone).
Among the baselines, ReSplat (Recurrent\,4) requires $\timingTotalresplatresplat$\,ms and VolSplat $\timingTotalvolsplatvolsplat$\,ms per scene.

\textbf{Storage size and rendering throughput.}
We additionally measure uncompressed storage size and rendering throughput using the same protocol as the main NVS evaluation (DA3 posed backbone, images resized so the longer side is 476\,px as in \secref{sec:eval_protocol}, 3/6/9/12 input views), computed on the 9 MipNeRF360 scenes.
\tabref{tab:timing_breakdown} reports the resulting Gaussian count, uncompressed \texttt{.ply} storage size, per-frame rendering time (at randomly sampled novel viewpoints, decoupled from the specific context/target view split), and the rendering-vs-reconstruction break-even point across all four view counts.

\begin{table*}[t]
\centering
\caption{Storage size and rendering throughput vs.\ input view count (DA3 posed backbone, MipNeRF360, same protocol as the main NVS evaluation).}
\vspace{0.2cm}
\label{tab:timing_breakdown}
\setlength{\tabcolsep}{3pt}
\scriptsize
\begin{tabular}{@{}r cc cc c cc c c c@{}}
\toprule
& \multicolumn{2}{c}{\#Gaussians (M)} & \multicolumn{2}{c}{Storage (MB)} & & \multicolumn{2}{c}{Render (ms/frame)} & & Merge & Break-even \\
\cmidrule(lr){2-3}\cmidrule(lr){4-5}\cmidrule(lr){7-8}
\#Views & Unmerged & Ours & Unmerged & Ours & Ratio$\downarrow$ & Unmerged & Ours & Speedup & overhead (ms) & (frames) \\
\midrule
3 & 0.51 & 0.02 & 125 & 5.5 & 22.9$\times$ & 1.74 & 0.71 & 2.4$\times$ & 86 & 84 \\
6 & 1.01 & 0.04 & 249 & 10.6 & 23.6$\times$ & 2.47 & 0.74 & 3.3$\times$ & 185 & 107 \\
9 & 1.52 & 0.06 & 374 & 16.1 & 23.5$\times$ & 3.15 & 0.77 & 4.1$\times$ & 304 & 128 \\
12 & 2.02 & 0.08 & 501 & 20.6 & 24.4$\times$ & 3.83 & 0.82 & 4.7$\times$ & 452 & 150 \\
\bottomrule

\end{tabular}
\end{table*}

At 12 input views, the unmerged per-pixel Gaussians require \storageUnmergedMBTwelve\,MB in uncompressed \texttt{.ply} format, compared to \storageMergedMBTwelve\,MB for our compacted $K\!=\!1$ output --- a \storageCompressionRatioTwelve$\times$ reduction.
Attribute-compression schemes such as HAC/HAC++~\cite{Chen2024HAC,hac++2025} are complementary to our primitive-count compaction and could be applied on top of our already-compacted output for further storage savings.

\begin{figure}[t]
\centering
\input{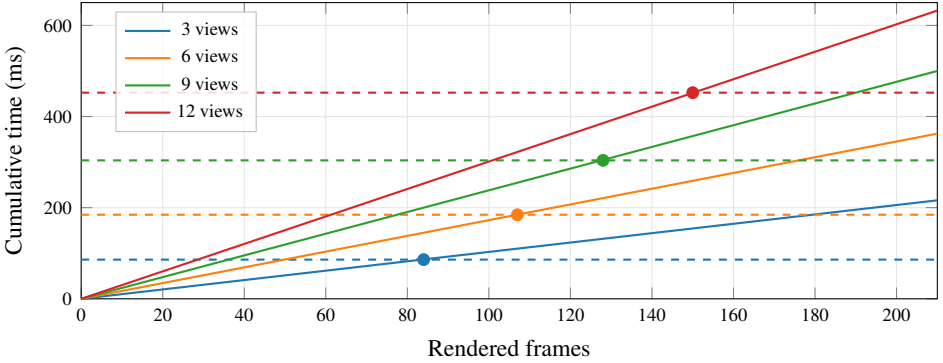}%
\definecolor{breakevenColThree}{HTML}{1F77B4}   %
\definecolor{breakevenColSix}{HTML}{FF7F0E}     %
\definecolor{breakevenColNine}{HTML}{2CA02C}    %
\definecolor{breakevenColTwelve}{HTML}{D62728}  %

\begin{tikzpicture}
\begin{axis}[
  width=\linewidth,
  height=5.5cm,
  xlabel={\small Rendered frames},
  ylabel={\small Cumulative time (ms)},
  xmin=0, xmax=\breakevenXmax,
  ymin=0, ymax=\breakevenYmax,
  grid=major,
  grid style={gray!20},
  tick label style={font=\scriptsize},
  label style={font=\small},
  legend style={
    font=\scriptsize,
    at={(0.04,0.97)},
    anchor=north west,
    draw=gray!50,
    fill=white,
    fill opacity=0.85,
    text opacity=1,
    row sep=0pt,
  },
  legend columns=1,
  clip=false,
]

\addplot[color=breakevenColThree, dashed, thick, forget plot]
  coordinates {\breakevenOverheadCoordsThree};
\addplot[color=breakevenColThree, thick]
  coordinates {\breakevenSavingsCoordsThree};
\addlegendentry{3 views}
\addplot[color=breakevenColThree, mark=*, mark size=2.2pt, only marks, forget plot]
  coordinates {\breakevenPointThree};

\addplot[color=breakevenColSix, dashed, thick, forget plot]
  coordinates {\breakevenOverheadCoordsSix};
\addplot[color=breakevenColSix, thick]
  coordinates {\breakevenSavingsCoordsSix};
\addlegendentry{6 views}
\addplot[color=breakevenColSix, mark=*, mark size=2.2pt, only marks, forget plot]
  coordinates {\breakevenPointSix};

\addplot[color=breakevenColNine, dashed, thick, forget plot]
  coordinates {\breakevenOverheadCoordsNine};
\addplot[color=breakevenColNine, thick]
  coordinates {\breakevenSavingsCoordsNine};
\addlegendentry{9 views}
\addplot[color=breakevenColNine, mark=*, mark size=2.2pt, only marks, forget plot]
  coordinates {\breakevenPointNine};

\addplot[color=breakevenColTwelve, dashed, thick, forget plot]
  coordinates {\breakevenOverheadCoordsTwelve};
\addplot[color=breakevenColTwelve, thick]
  coordinates {\breakevenSavingsCoordsTwelve};
\addlegendentry{12 views}
\addplot[color=breakevenColTwelve, mark=*, mark size=2.2pt, only marks, forget plot]
  coordinates {\breakevenPointTwelve};

\end{axis}
\end{tikzpicture}
\caption{Rendering-vs-reconstruction break-even (DA3 posed backbone, MipNeRF360) for 3/6/9/12 input views.
Dashed lines: constant one-time merging overhead.
Solid lines: cumulative rendering time saved by our compacted $K\!=\!1$ output vs.\ the unmerged backbone output.
Markers indicate the break-even frame count per view count.}
\label{fig:breakeven}
\end{figure}

\textbf{Rendering-vs-reconstruction break-even.}
Rendering our compacted $K\!=\!1$ Gaussians is \renderSpeedupTwelve$\times$ faster per frame than the unmerged backbone output at 12 input views (\renderMsMergedTwelve\,ms vs.\ \renderMsUnmergedTwelve\,ms per frame).
Our merging pipeline (superpixel segmentation, encoding, cross-view matching and merging, and decoding) adds \mergeOverheadMsTwelve\,ms of reconstruction overhead over direct backbone inference at 12 views.
Dividing this overhead by the per-frame rendering time saved gives a break-even point of \breakEvenFramesTwelve\ rendered frames, after which the cumulative rendering-time savings exceed the added reconstruction cost; this ranges from \breakEvenFramesThree\ frames at 3 input views to \breakEvenFramesTwelve\ frames at 12 (\tabref{tab:timing_breakdown}).
\figref{fig:breakeven} visualizes this trade-off directly: for each view count, the dashed line marks the (constant) one-time reconstruction overhead, and the solid line marks the cumulative rendering time saved as a function of rendered frames; their intersection is the break-even point.
For applications rendering many frames per reconstructed scene at high resolution (\eg, simulation for autonomous driving or robotics), the amortized rendering throughput benefit of compaction typically dominates the one-time reconstruction overhead.

\section{Decoder Diversification Regularizer}
\label{sec:diversification}

For $K\!>\!1$ decoder heads, a regularizer prevents the $K$ output Gaussians from collapsing into identical copies while encouraging spatial coverage of the original superpixel region.
Both terms below are pair-size-weighted: each pair contribution is scaled by $w_{ij} = \sqrt{v_i v_j}$ where $v_i = \exp(\sum_k s_{k,i})$ is the volume proxy of Gaussian~$i$.

For each pair $(i, j)$ within a group, we evaluate a U-shaped profile $f$ over the AABB IoU $u_{ij}$ and the normalized gap $\gamma_{ij} = \max(0,\, d_{ij} - d^\text{contact}_{ij}) / d^\text{contact}_{ij}$, where $d_{ij}$ is the Euclidean distance between means and $d^\text{contact}_{ij}$ is the sum of the two AABB half-extent norms.
The profile switches between a repulsion branch (IoU $\geq 0.5$) and an attraction branch (IoU $< 0.5$):
\begin{equation}
f(u_{ij}, \gamma_{ij}) =
\begin{cases}
\left(\dfrac{u_{ij} - 0.5}{0.5}\right)^{\!3} & u_{ij} \geq 0.5, \\[8pt]
0.5\!\left(\dfrac{0.5 - u_{ij}}{0.5}\right)^{\!3} + 0.5\!\left(1 - e^{-3\gamma_{ij}}\right) & u_{ij} < 0.5.
\end{cases}
\end{equation}
The repulsion branch penalizes excessive overlap; the attraction branch prevents Gaussians from drifting apart when non-overlapping, with the distance term saturating at a gap of twice the contact distance.
The geometric term averages these weighted profile values over all pairs $\mathcal{P}$:
\begin{equation}
\mathcal{L}_\text{geom} = \frac{1}{|\mathcal{P}|} \sum_{(i,j)\in\mathcal{P}} w_{ij}\, f(u_{ij}, \gamma_{ij}).
\end{equation}
To prevent single-Gaussian dominance, we penalize pairwise opacity differences:
\begin{equation}
\mathcal{L}_\text{opa} = \frac{1}{|\mathcal{P}|} \sum_{(i,j)\in\mathcal{P}} w_{ij}\, \lvert\alpha_i - \alpha_j\rvert^2.
\end{equation}

The combined diversification loss blends both terms:
\begin{equation}
\mathcal{L}_\text{div} = 0.75\,\mathcal{L}_\text{geom} + 0.25\,\mathcal{L}_\text{opa}.
\end{equation}

\begin{table}[t]
\centering
\caption{Loss weights and decay schedules.}
\vspace{0.2cm}
\label{tab:hyperparams}
\scriptsize
\begin{tabular}{@{}lccl@{}}
\toprule
Loss term & Weight & Decay schedule & Notes \\
\midrule
$\mathcal{L}_\text{MSE}$ & 1.0 & -- & MSE \\
$\mathcal{L}_\text{SSIM}$ & 0.05 & -- & SSIM \\
$\mathcal{L}_\text{LPIPS}$ & 0.05 & -- & LPIPS with VGG backbone \\
$\mathcal{L}_\text{teach}$ & 0.25 & $\to 0$ at 2k steps & Moment-matching teacher \\
$\mathcal{L}_\text{div}$ & 1.0 & $\to 0$ at 50k steps & Decoder diversification (attraction-repulsion) \\
\bottomrule
\end{tabular}
\end{table}

\section{Bounding the Compute Budget of the Encoder}
\label{sec:encoder_compute}

In order to allow maximum compression capability, we do not impose a hard limit on the number of Gaussians that can be grouped into a superpixel, and thus encoded into a single Feature Gaussian. 
For textureless regions, the superpixel segmentation can produce very large segments that group hundreds or even thousands of Gaussians, which could lead to an excessively high computational cost during encoding due to the attention mechanism.
We therefore subsample a fixed number of Gaussians from each superpixel group exceeding an absolute upper limit, which allows us to maintain a bounded compute budget regardless of the superpixel size. 
To allow the representation to still encode the visual importance of the entire group, we compute the group mass $m$ across all member Gaussians (not just the subsampled ones), append it to the parameters of the Feature Gaussian, and provide it as an additional input to all modules.

The group mass sums the individual Gaussian contributions:
\begin{equation}
\label{eq:mass}
m = \sum_{i=1}^{n} \alpha_i\,(2\pi)^{3/2}\,\exp(s_{x,i} + s_{y,i} + s_{z,i}),
\end{equation}
reflecting each primitive's opacity-weighted volume.

Additionally, we bin the superpixel groups according to their number of member Gaussians, and pad the input to the encoder to the maximum group size within each bin, allowing to parallelize the encoding of groups with similar sizes while still supporting a wide range of group sizes.

\section{Input-View Reconstruction Quality}
\label{sec:recon_tables}

\begin{table*}[t]
\centering
\caption{Input-view reconstruction quality for the same backbone variants and baselines as in\,\tabref{tab:nvs}.
Best in \textbf{bold}, second best \underline{underlined}.}
\vspace{0.2cm}
\label{tab:recon_all}
\setlength{\tabcolsep}{2.2pt}
\scriptsize
\begin{tabular}{@{}ll r ccc ccc ccc@{}}
\toprule
& & & \multicolumn{3}{c}{DL3DV-Bench} & \multicolumn{3}{c}{MipNeRF360} & \multicolumn{3}{c}{T\&T} \\
\cmidrule(lr){4-6}\cmidrule(lr){7-9}\cmidrule(lr){10-12}
Backbone & Method & $r_c$\,(\%)$\downarrow$ & PSNR$\uparrow$ & SSIM$\uparrow$ & LPIPS$\downarrow$ & PSNR$\uparrow$ & SSIM$\uparrow$ & LPIPS$\downarrow$ & PSNR$\uparrow$ & SSIM$\uparrow$ & LPIPS$\downarrow$ \\
\midrule
\multirow{6}{*}{AnySplat} & Unmerged             &    100.0\% &                 15.78 &                 0.407 &                 0.428 &                 18.58 &                 0.449 &     \underline{0.364} &                 16.50 &                 0.487 &                 0.409 \\
 & Voxelized            &     83.2\% &                 15.84 &                 0.406 &                 0.433 &                 18.62 &                 0.449 &                 0.365 &                 16.54 &                 0.482 &                 0.422 \\
\cmidrule{2-12}
 & Mom.\ Match.         &      5.3\% &                 15.91 &                 0.413 &                 0.564 &                 16.73 &                 0.387 &                 0.590 &                 15.60 &                 0.469 &                 0.555 \\
 & Ours ($K\!=\!1$)     & \underline{4.4\%} &                 15.38 &                 0.392 &                 0.522 &                 17.83 &                 0.419 &                 0.529 &                 15.75 &                 0.448 &                 0.517 \\
 & Ours ($K\!=\!2$)     &      8.9\% &                 15.56 &                 0.398 &                 0.517 &                 18.17 &                 0.427 &                 0.524 &                 16.15 &                 0.464 &                 0.510 \\
 & Ours ($K\!=\!4$)     &     17.8\% &                 15.65 &                 0.402 &                 0.514 &                 18.36 &                 0.431 &                 0.522 &                 16.27 &                 0.469 &                 0.507 \\
\midrule
\multirow{5}{*}{DepthSplat} & Unmerged             &    100.0\% &                 17.11 &        \textbf{0.652} &     \underline{0.300} &                 16.49 &                 0.534 &                 0.368 &                 16.16 &     \underline{0.633} &        \textbf{0.343} \\
\cmidrule{2-12}
 & Mom.\ Match.         &      6.1\% &                 13.19 &                 0.446 &                 0.609 &                 12.23 &                 0.341 &                 0.685 &                 11.68 &                 0.450 &                 0.643 \\
 & Ours ($K\!=\!1$)     &      5.3\% &                 18.41 &                 0.578 &                 0.433 &                 17.97 &                 0.486 &                 0.492 &                 16.94 &                 0.558 &                 0.484 \\
 & Ours ($K\!=\!2$)     &     10.5\% &                 18.61 &                 0.583 &                 0.427 &                 18.68 &                 0.501 &                 0.482 &                 17.26 &                 0.568 &                 0.479 \\
 & Ours ($K\!=\!4$)     &     21.1\% &                 18.72 &                 0.587 &                 0.424 &                 18.99 &                 0.508 &                 0.477 &                 17.41 &                 0.572 &                 0.477 \\
\midrule
\multirow{5}{*}{DA3} & Unmerged             &    100.0\% &                 18.84 &                 0.612 &        \textbf{0.281} &                 18.23 &                 0.439 &                 0.367 &                 16.48 &                 0.521 &     \underline{0.366} \\
\cmidrule{2-12}
 & Mom.\ Match.         &      4.9\% &                 13.74 &                 0.426 &                 0.696 &                 13.88 &                 0.344 &                 0.737 &                 12.45 &                 0.438 &                 0.736 \\
 & Ours ($K\!=\!1$)     & \textbf{4.4\%} &                 19.21 &                 0.579 &                 0.431 &                 19.38 &                 0.466 &                 0.490 &                 17.74 &                 0.537 &                 0.472 \\
 & Ours ($K\!=\!2$)     &      8.8\% &     \underline{19.32} &                 0.585 &                 0.424 &                 19.47 &                 0.470 &                 0.485 &     \underline{17.85} &                 0.541 &                 0.466 \\
 & Ours ($K\!=\!4$)     &     17.3\% &        \textbf{19.37} &                 0.586 &                 0.421 &                 19.51 &                 0.470 &                 0.482 &                 17.84 &                 0.541 &                 0.464 \\
\midrule
\midrule
\multirow{2}{*}{ReSplat} & Init                 &      6.2\% &                 14.67 &                 0.509 &                 0.462 &     \underline{19.70} &     \underline{0.626} &                 0.395 &                 15.89 &                 0.593 &                 0.428 \\
 & Recurrent~4          &      6.2\% &                 17.77 &     \underline{0.633} &                 0.389 &        \textbf{22.45} &        \textbf{0.687} &        \textbf{0.335} &        \textbf{18.03} &        \textbf{0.665} &                 0.372 \\
VolSplat & ---                  &     93.4\% &                 16.19 &                 0.518 &                 0.522 &                 13.47 &                 0.383 &                 0.618 &                 14.34 &                 0.512 &                 0.519 \\
\bottomrule

\end{tabular}
\end{table*}

\tabref{tab:recon_all} reports the input-view reconstruction quality for all backbone variants.
ReSplat generally achieves the best input reconstruction quality, as it iteratively refines the primitives based on the rendering error.
As shown by the novel view synthesis results in the main paper, this does not translate to better novel view quality, as the refinement is prone to overfitting to the input views.

\section{Per-View-Count NVS Quality}
\label{sec:per_view}

The main paper reports NVS metrics averaged over all four input-view counts (3, 6, 9, 12 views) to give a single summary number.
\tabref{tab:per_view} breaks this down by view count for each of the three backbone variants from the main table.
Notice that the unseen regions in the GT views are masked, so the metrics do not increase dramatically with more views as they would if the entire GT image was considered.
ReSplat performs much worse with fewer views, as the refinement is more prone to overfitting to the sparse input views and thus does not generalize as well to novel views.
Meanwhile, our approach maintains a more consistent performance across view counts, leveraging the strong sparse-view capabilities of the feed-forward backbone while improving the performance in terms of PSNR and SSIM through the merging and decoding process.

\begin{table*}[t]
\centering
\caption{Novel view synthesis quality per input view count, averaged across DL3DV-Bench, MipNeRF360, and T\&T.
Best in \textbf{bold}, second best \underline{underlined}.}
\vspace{0.2cm}
\label{tab:per_view}
\setlength{\tabcolsep}{1.5pt}
\scriptsize
\begin{tabular}{@{}ll ccc ccc ccc ccc@{}}
\toprule
& & \multicolumn{3}{c}{3 views} & \multicolumn{3}{c}{6 views} & \multicolumn{3}{c}{9 views} & \multicolumn{3}{c}{12 views} \\
\cmidrule(lr){3-5}\cmidrule(lr){6-8}\cmidrule(lr){9-11}\cmidrule(lr){12-14}
Backbone & Method & PSNR$\uparrow$ & SSIM$\uparrow$ & LPIPS$\downarrow$ & PSNR$\uparrow$ & SSIM$\uparrow$ & LPIPS$\downarrow$ & PSNR$\uparrow$ & SSIM$\uparrow$ & LPIPS$\downarrow$ & PSNR$\uparrow$ & SSIM$\uparrow$ & LPIPS$\downarrow$ \\
\midrule
\multirow{6}{*}{AnySplat} & Unmerged             &         12.07 &         0.392 &         0.471 &         12.37 &         0.379 &         0.475 &         12.81 &         0.366 &         0.494 &         13.13 &         0.358 &         0.505 \\
 & Voxelized            &         12.05 &         0.389 &         0.473 &         12.39 &         0.378 &         0.477 &         12.83 &         0.365 &         0.495 &         13.17 &         0.357 &         0.506 \\
\cmidrule{2-14}
 & Mom.\ Match.         &         12.71 &         0.424 &         0.527 &         13.41 &         0.434 &         0.534 &         13.80 &         0.423 &         0.556 &         13.94 &         0.413 &         0.572 \\
 & Ours ($K\!=\!1$)     &         13.24 &         0.421 &         0.526 &         14.11 &         0.435 &         0.521 &         14.51 &         0.423 &         0.537 &         14.70 &         0.414 &         0.547 \\
 & Ours ($K\!=\!2$)     &         13.89 &         0.434 &         0.528 &         14.47 &         0.441 &         0.520 &         14.66 &         0.427 &         0.536 &         14.80 &         0.417 &         0.547 \\
 & Ours ($K\!=\!4$)     &         13.94 &         0.437 &         0.526 &         14.48 &         0.440 &         0.518 &         14.71 &         0.427 &         0.534 &         14.88 &         0.418 &         0.545 \\
\midrule
\multirow{5}{*}{DepthSplat} & Unmerged             &         15.11 &  \textbf{0.584} &  \underline{0.318} &         14.05 &         0.537 &  \underline{0.362} &         14.23 &         0.514 &  \underline{0.393} &         14.56 &         0.506 &  \underline{0.411} \\
\cmidrule{2-14}
 & Mom.\ Match.         &         12.03 &         0.478 &         0.537 &         12.08 &         0.470 &         0.545 &         12.46 &         0.456 &         0.576 &         12.68 &         0.448 &         0.600 \\
 & Ours ($K\!=\!1$)     &         15.85 &         0.533 &         0.423 &         15.23 &         0.501 &         0.445 &         15.57 &         0.482 &         0.463 &         15.86 &         0.475 &         0.472 \\
 & Ours ($K\!=\!2$)     &         16.30 &         0.546 &         0.420 &         15.72 &         0.512 &         0.441 &         15.90 &         0.490 &         0.460 &         16.15 &         0.482 &         0.469 \\
 & Ours ($K\!=\!4$)     &         16.51 &         0.551 &         0.418 &         15.90 &         0.516 &         0.439 &         16.09 &         0.494 &         0.458 &         16.29 &         0.485 &         0.467 \\
\midrule
\multirow{5}{*}{DA3} & Unmerged             &         16.77 &         0.576 &  \textbf{0.306} &         16.42 &         0.552 &  \textbf{0.340} &         16.90 &         0.546 &  \textbf{0.346} &         17.18 &  \textbf{0.540} &  \textbf{0.351} \\
\cmidrule{2-14}
 & Mom.\ Match.         &         13.52 &         0.492 &         0.574 &         13.46 &         0.477 &         0.603 &         13.43 &         0.458 &         0.634 &         13.41 &         0.445 &         0.654 \\
 & Ours ($K\!=\!1$)     &         17.58 &         0.575 &         0.404 &         17.07 &         0.554 &         0.425 &         17.41 &         0.543 &         0.439 &         17.58 &         0.534 &         0.446 \\
 & Ours ($K\!=\!2$)     &  \underline{17.75} &         0.580 &         0.400 &  \underline{17.19} &  \underline{0.558} &         0.422 &  \underline{17.54} &  \underline{0.546} &         0.436 &  \underline{17.67} &         0.537 &         0.443 \\
 & Ours ($K\!=\!4$)     &  \textbf{17.77} &  \underline{0.581} &         0.400 &  \textbf{17.23} &  \textbf{0.559} &         0.421 &  \textbf{17.57} &  \textbf{0.547} &         0.435 &  \textbf{17.69} &  \underline{0.537} &         0.442 \\
\midrule
\midrule
\multirow{2}{*}{ReSplat} & Init                 &         12.37 &         0.323 &         0.598 &         13.15 &         0.358 &         0.587 &         14.23 &         0.403 &         0.557 &         15.05 &         0.425 &         0.542 \\
 & Recurrent~4          &         14.06 &         0.409 &         0.582 &         14.59 &         0.422 &         0.571 &         15.70 &         0.457 &         0.535 &         16.41 &         0.476 &         0.517 \\
VolSplat & ---                  &         12.63 &         0.302 &         0.617 &         12.09 &         0.307 &         0.665 &         12.80 &         0.320 &         0.662 &         12.65 &         0.322 &         0.665 \\
\bottomrule

\end{tabular}
\end{table*}

\section{Method extensions}
\label{sec:extensions}

\textbf{Content-Adaptive Per-Superpixel $K$.}
The content-adaptive distribution of primitives in our pipeline is handled entirely by the saliency-guided superpixel segmentation (\secref{sec:grouping}), which already allocates larger superpixels to homogeneous regions and smaller ones to detailed areas; the $K$ decoder heads instead offer a single global knob for the overall quality--efficiency trade-off (\secref{sec:decoder}).
As an alternative, training-free mechanism for content-adaptivity, we evaluate assigning a mixed, per-superpixel $K$ based on the saliency map $\boldsymbol{\lambda}_v$, instead of a fixed global $K$: each superpixel is assigned to one of the three decoder heads ($K\!\in\!\{1,2,4\}$) by its saliency percentile, so that each head decodes one third of the superpixels, ranked from least to most salient.
With our segmentation, this mixed-$K$ variant (\psnrdaThreemergedkmixed\,dB at \rcdaThreemergedkmixed\% $r_c$) does not yield a favorable trade-off compared to simply using a larger fixed $K$: it reaches a similar quality--efficiency point as the $K\!=\!2$ decoder alone (\psnrdaThreemergedktwoFlexK\,dB at \rcdaThreemergedktwoFlexK\% $r_c$).
This suggests that, for a segmentation method that already places larger segments in flat regions and smaller ones at detailed boundaries, allocating additional primitives per superpixel based on saliency does not provide further benefits.
A per-superpixel $K$ mechanism may however be beneficial for less content-adaptive segmentation methods that produce more uniformly sized superpixels.

\textbf{Compaction of Non-Pixel-Aligned FF Backbones.}
Our grouping stage assumes a one-to-one correspondence between image pixels and Gaussians to build the partition map $\mathbf{M}_v$ (\secref{sec:grouping}).
Backbones that instead predict voxel-aligned or otherwise non-pixel-aligned Gaussians (\eg, VolSplat~\cite{wang2025volsplat} or AnySplat's~\cite{Jiang2025} built-in voxelization) do not admit this mapping directly.
However, their already-reduced primitive sets still leave room for further compaction (\eg, AnySplat's voxelization retains \rcanysplatVarFeatvoxelized\% of the original count, \tabref{tab:nvs}).
Supporting such backbones would let our pipeline compound the efficiency already gained during reconstruction and offer the same controllable level-of-detail on top of a more compact starting representation.

To recover a per-Gaussian superpixel grouping without pixel alignment, we compute the rendering contribution of every Gaussian to every superpixel according to the rasterization described in \secref{sec:prelim}.
For every (Gaussian, pixel) pair visited by the renderer, we compute the alpha-compositing weight $w_{ip} = \alpha_{ip}\,T_{ip}$ in closed form.
Summing these weights per (Gaussian, superpixel) pair with a sparse tensor, each Gaussian is assigned to its dominant superpixel:
\begin{equation}
s_i^\star = \argmax_s \sum_{p \in \mathcal{S}_s} w_{ip}
\end{equation}
Gaussians with negligible accumulated weight across all views (occluded or outside all view frustums) are marked invalid and excluded from the final set.
This recovers a dominant view and superpixel id for every valid Gaussian, after which our encoding, cross-view matching and merging, and decoding stages (\secref{sec:encoder}--\secref{sec:merging}) apply unchanged.

\begin{table}[t]
\centering
\caption{Compaction of AnySplat's voxelized output vs.\ merging its per-pixel Gaussians directly, averaged over DL3DV-Bench, MipNeRF360, and T\&T (all four input view counts).
Best in \textbf{bold}, second best \underline{underlined}.}
\vspace{0.2cm}
\label{tab:voxel_compaction}
\scriptsize
\begin{tabular}{@{}l r ccc@{}}
\toprule
Method & $r_c$\,(\%)$\downarrow$ & PSNR$\uparrow$ & SSIM$\uparrow$ & LPIPS$\downarrow$ \\
\midrule
AnySplat (unmerged) & 100.00 & 12.68 & 0.3784 & \textbf{0.4860} \\
AnySplat (voxelized) & 83.76 & 12.69 & 0.3768 & \underline{0.4872} \\
\midrule
Merged, per-pixel ($K\!=\!1$) & \underline{4.20} & 14.22 & 0.4322 & 0.5312 \\
Merged, per-pixel ($K\!=\!2$) & 8.40 & 14.52 & \textbf{0.4371} & 0.5321 \\
Merged, per-pixel ($K\!=\!4$) & 16.78 & \underline{14.53} & 0.4362 & 0.5301 \\
\midrule
Merged, voxelized ($K\!=\!1$) & \textbf{3.58} & 14.09 & 0.4241 & 0.5278 \\
Merged, voxelized ($K\!=\!2$) & 7.15 & 14.43 & 0.4317 & 0.5266 \\
Merged, voxelized ($K\!=\!4$) & 14.28 & \textbf{14.58} & \underline{0.4363} & 0.5241 \\
\bottomrule
\end{tabular}
\end{table}

\tabref{tab:voxel_compaction} reports results for applying our pipeline to AnySplat's voxelized output, in place of the per-pixel Gaussians, using the same benchmarks and metrics as \tabref{tab:nvs}.
Merging the already-compact voxelized representation compounds both compaction stages: the $K\!=\!1$ decoder reaches $r_c\!=\!3.58\%$, the lowest primitive count of all variants.
All three metrics (PSNR, SSIM, LPIPS) show a similar performance, whether applied to the per-pixel or voxelized backbones, with a slight drop in PSNR and SSIM for the voxelized variant for the $K=1$ and $K=2$ decoders.
LPIPS degrades similarly for both variants relative to the unmerged and voxelized backbones, consistent with the general perceptual-detail loss from merging discussed in \secref{sec:conclusion}.
These results show that while our pipeline is optimized for and works best with pixel-aligned backbones, it can also be applied on top of voxel-based backbones such as AnySplat's built-in voxelization, further compounding their efficiency gains.

\section{Additional Qualitative Results}
\label{sec:qualitative_suppl}

\figref{fig:qualitative_suppl_dl3dv}, \figref{fig:qualitative_suppl_mipnerf360}, and \figref{fig:qualitative_suppl_tnt} extend the qualitative comparison from the main paper (\figref{fig:qualitative}) to additional scenes from all three benchmarks, using the same DA3 backbone.
Columns show the ground truth and backbone reconstructions, alongside our decoder output at $K\!=\!1$ and both baselines.

\section{Posed vs.\ Unposed DA3 Backbone}
\label{sec:da3_posed_unposed}

\begin{table*}[t]
\centering
\caption{Novel view synthesis quality for posed and unposed DA3 backbones, reported per benchmark (averaged over 3, 6, 9, and 12 input views).
Best in \textbf{bold}, second best \underline{underlined}.}
\vspace{0.2cm}
\label{tab:da3_posed_unposed}
\setlength{\tabcolsep}{2.2pt}
\scriptsize
\begin{tabular}{@{}ll r ccc ccc ccc@{}}
\toprule
& & & \multicolumn{3}{c}{DL3DV-Bench} & \multicolumn{3}{c}{MipNeRF360} & \multicolumn{3}{c}{T\&T} \\
\cmidrule(lr){4-6}\cmidrule(lr){7-9}\cmidrule(lr){10-12}
Backbone & Method & $r_c$\,(\%)$\downarrow$ & PSNR$\uparrow$ & SSIM$\uparrow$ & LPIPS$\downarrow$ & PSNR$\uparrow$ & SSIM$\uparrow$ & LPIPS$\downarrow$ & PSNR$\uparrow$ & SSIM$\uparrow$ & LPIPS$\downarrow$ \\
\midrule
\multirow{5}{*}{DA3} & Unmerged             &    100.0\% &        \textbf{18.68} &        \textbf{0.620} &        \textbf{0.297} &                 16.71 &                 0.495 &        \textbf{0.356} &                 15.07 &                 0.546 &        \textbf{0.353} \\
\cmidrule{2-12}
 & Mom.\ Match.         &      4.9\% &                 13.79 &                 0.459 &                 0.637 &                 14.04 &                 0.440 &                 0.601 &                 12.55 &                 0.505 &                 0.611 \\
 & Ours ($K\!=\!1$)     & \underline{4.4\%} &                 18.37 &                 0.578 &                 0.422 &                 17.87 &                 0.514 &                 0.442 &                 16.00 &                 0.562 &                 0.421 \\
 & Ours ($K\!=\!2$)     &      8.8\% &                 18.50 &                 0.582 &                 0.417 &     \underline{18.02} &     \underline{0.517} &                 0.440 &     \underline{16.09} &     \underline{0.566} &                 0.418 \\
 & Ours ($K\!=\!4$)     &     17.3\% &     \underline{18.54} &     \underline{0.583} &                 0.416 &        \textbf{18.06} &        \textbf{0.518} &                 0.439 &        \textbf{16.09} &        \textbf{0.566} &                 0.418 \\
\midrule
\multirow{5}{*}{DA3 (Unposed)} & Unmerged             &    100.0\% &                 18.25 &                 0.573 &     \underline{0.315} &                 16.50 &                 0.464 &     \underline{0.381} &                 14.67 &                 0.507 &     \underline{0.398} \\
\cmidrule{2-12}
 & Mom.\ Match.         &      4.9\% &                 13.69 &                 0.454 &                 0.644 &                 14.00 &                 0.434 &                 0.607 &                 12.30 &                 0.498 &                 0.620 \\
 & Ours ($K\!=\!1$)     & \textbf{4.3\%} &                 17.92 &                 0.546 &                 0.429 &                 17.36 &                 0.488 &                 0.464 &                 15.19 &                 0.526 &                 0.453 \\
 & Ours ($K\!=\!2$)     &      8.6\% &                 17.99 &                 0.550 &                 0.424 &                 17.47 &                 0.492 &                 0.459 &                 15.22 &                 0.528 &                 0.450 \\
 & Ours ($K\!=\!4$)     &     17.1\% &                 18.02 &                 0.551 &                 0.424 &                 17.49 &                 0.493 &                 0.458 &                 15.23 &                 0.529 &                 0.450 \\
\bottomrule

\end{tabular}
\end{table*}

The DA3 backbone can be used in a posed or unposed configuration.
In the main paper we report the posed variant, which achieves better NVS quality by leveraging the explicit camera information to predict a more view-consistent representation.
This is consistent with the used baselines, which both rely on camera poses.
In \tabref{tab:da3_posed_unposed}, we compare the posed and unposed DA3 variants across all three benchmarks, showing that even without pose conditioning, our approach still outperforms the baselines.

\clearpage

\begin{figure}[p]
\centering
\setlength{\qualimw}{0.192\textwidth}
\setlength{\tabcolsep}{1pt}
\begin{tabular}{@{}ccccc@{}}
\toprule
\input{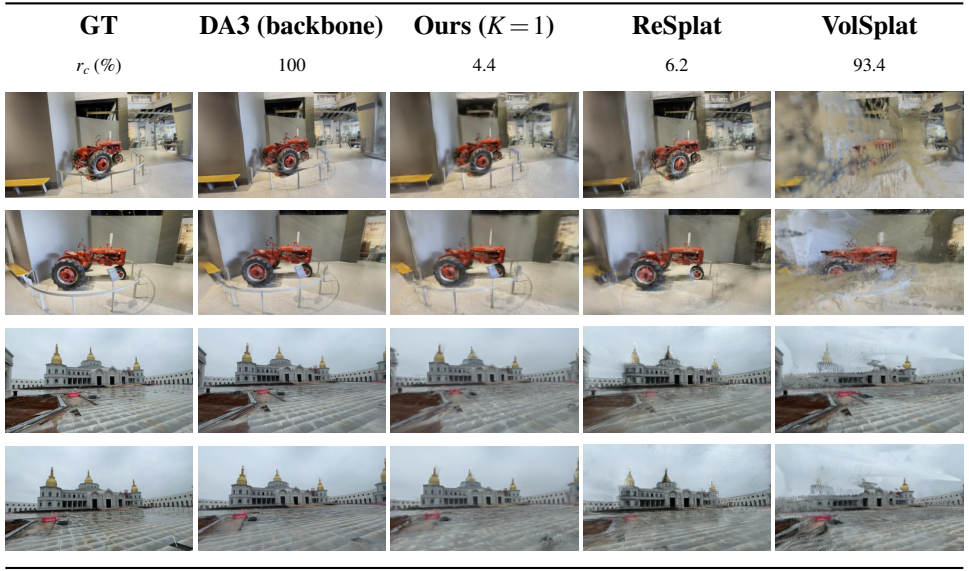}
\end{tabular}
\caption{Additional qualitative results on DL3DV-Bench (12 input views, DA3 backbone).}
\label{fig:qualitative_suppl_dl3dv}
\end{figure}

\begin{figure}[p]
\centering
\setlength{\qualimw}{0.192\textwidth}
\setlength{\tabcolsep}{1pt}
\begin{tabular}{@{}ccccc@{}}
\toprule
\input{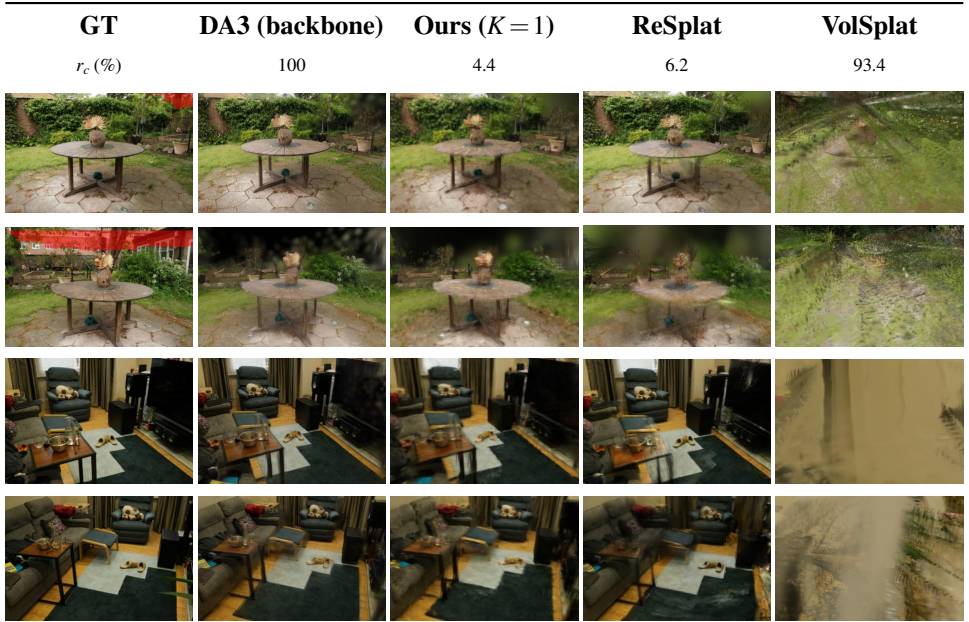}
\end{tabular}
\caption{Additional qualitative results on MipNeRF360 (12 input views, DA3 backbone).}
\label{fig:qualitative_suppl_mipnerf360}
\end{figure}

\begin{figure}[p]
\centering
\setlength{\qualimw}{0.192\textwidth}
\setlength{\tabcolsep}{1pt}
\begin{tabular}{@{}ccccc@{}}
\toprule
\input{pics/gen/qualitative_suppl_tanks_and_temples}
\end{tabular}
\caption{Additional qualitative results on Tanks \& Temples (12 input views, DA3 backbone).}
\label{fig:qualitative_suppl_tnt}
\end{figure}

\end{document}